\documentclass[10pt,twocolumn,letterpaper]{article}

\usepackage{cvpr}              

\usepackage{graphicx}
\usepackage{algorithm}
\usepackage{algorithmic}
\usepackage[algo2e,linesnumbered,noend]{algorithm2e}
\usepackage{amsmath,bm}
\usepackage{amssymb}
\usepackage{multirow}
\usepackage{booktabs}
\usepackage{color}
\usepackage{float}
\usepackage{tabularx}
\usepackage[T1]{fontenc}
\usepackage[accsupp]{axessibility} 
\DeclareMathOperator*{\argmin}{\arg\min} 

\usepackage{color, colortbl}
\definecolor{orange}{rgb}{0.988, 0.878, 0.533}

\usepackage{enumitem}

\usepackage[pagebackref,breaklinks,colorlinks]{hyperref}

\usepackage{mathtools}

\DeclarePairedDelimiterX{\infdivx}[2]{(}{)}{%
  #1\;\delimsize\|\;#2%
}

\usepackage[capitalize]{cleveref}
\crefname{section}{Sec.}{Secs.}
\Crefname{section}{Section}{Sections}
\Crefname{table}{Table}{Tables}
\crefname{table}{Tab.}{Tabs.}

\def\cvprPaperID{5638} 
\def\confName{CVPR}
\def\confYear{2024}

\makeatletter
\DeclareRobustCommand\onedot{\futurelet\@let@token\@onedot}
\def\@onedot{\ifx\@let@token.\else.\null\fi\xspace}
\def\eg{\emph{e.g}\onedot} 
\def\ie{\emph{i.e}\onedot}

\def\etal{\emph{et al}\onedot}
\makeatother

\DeclareRobustCommand*{\IEEEauthorrefmark}[1]{%
    \raisebox{0pt}[0pt][0pt]{\textsuperscript{\footnotesize\ensuremath{#1}}}}

\newcommand{\mathbbm}[1]{\text{\usefont{U}{bbm}{m}{n}#1}}
\newcommand{\red}[1]{\textcolor{red}{#1}}

\newcommand{\appendixtitle}{
    \clearpage
    \twocolumn[
        \centering
        \Large
        \textbf{Supplementary Material}\\
        \vspace{0.5em}
    ]
    \newpage
}

\begin{document}

\title{Learning to Rematch Mismatched Pairs for Robust Cross-Modal Retrieval}
\author{Haochen Han\IEEEauthorrefmark1,
Qinghua Zheng\IEEEauthorrefmark1,
Guang Dai\IEEEauthorrefmark2, 
Minnan Luo\IEEEauthorrefmark{1}\thanks{Corresponding author. This work was supported by the National Key Research and Development Program of China (2022YFB3102600), and National Nature Science Foundation of China (62272374, 62192781, 62202367, 62250009, and 62137002).}\ ,
Jingdong Wang\IEEEauthorrefmark3
\and
{ \IEEEauthorrefmark1 Xi’an Jiaotong University}
{ \IEEEauthorrefmark2 SGIT AI Lab}
{ \IEEEauthorrefmark3 Baidu Inc}
\and
\texttt{\normalsize hhc1997@stu.xjtu.edu.cn, \{qhzheng, minnluo\}@xjtu.edu.cn}
\and
\texttt{\normalsize guang.gdai@gmail.com, wangjingdong@baidu.com}}

\maketitle

\begin{abstract}
Collecting well-matched multimedia datasets is crucial for training cross-modal retrieval models. However, in real-world scenarios, massive multimodal data are harvested from the Internet, which inevitably contains Partially Mismatched Pairs (PMPs). Undoubtedly, such semantical irrelevant data will remarkably harm the cross-modal retrieval performance. Previous efforts tend to mitigate this problem by estimating a soft correspondence to down-weight the contribution of PMPs. In this paper, we aim to address this challenge from a new perspective: the potential semantic similarity among unpaired samples makes it possible to excavate useful knowledge from mismatched pairs. To achieve this, we propose L2RM, a general framework based on Optimal Transport (OT) that learns to rematch mismatched pairs. In detail, L2RM aims to generate refined alignments by seeking a minimal-cost transport plan across different modalities. To formalize the rematching idea in OT, first, we propose a self-supervised cost function that automatically learns from explicit similarity-cost mapping relation. Second, we present to model a partial OT problem while restricting the transport among false positives to further boost refined alignments. Extensive experiments on three benchmarks demonstrate our L2RM significantly improves the robustness against PMPs for existing models. The code is available at \url{https://github.com/hhc1997/L2RM}.

\end{abstract}

\section{Introduction}
The pursuit of general intelligence has advanced the progress of multimodal learning, which aims to understand and integrate multiple sensory modalities like humans. Cross-modal retrieval is one of the most important techniques in multimodal learning due to its flexibility in bridging different modalities \cite{sheng2021human,huang2023vop,wang2023multilateral,hao2023dual,feng2023rono,jiang2023cross}, which has powered various real-world applications.
  
Despite the remarkable performance of previous methods, much of their success can be attributed to the voracious appetite for well-matched cross-modal pairs. In practice, collecting such ideal data \cite{jia2021scaling} is notoriously labor-intensive and even impossible. Alternatively, several mainstream cross-modal datasets utilize the co-occurred information to crawl data from the Internet, especially for visual-text samples \cite{desai2021redcaps}. Although such a data collection way is free from expensive annotations, it will inevitably introduce partially mismatched pairs. For example, the standard image-caption dataset, Conceptual Captions \cite{sharma2018conceptual}, is estimated to contain about 3\% to 20\% mismatched pairs. Such semantically irrelevant data will be wrongly treated as the matched pairs for training, which undoubtedly impairs the performance of cross-modal retrieval models. Thus, endowing cross-modal learning with robustness against PMPs is crucial to suit real-world retrieval scenarios.

\begin{figure}[t]
\centering  
\includegraphics[width=\columnwidth]{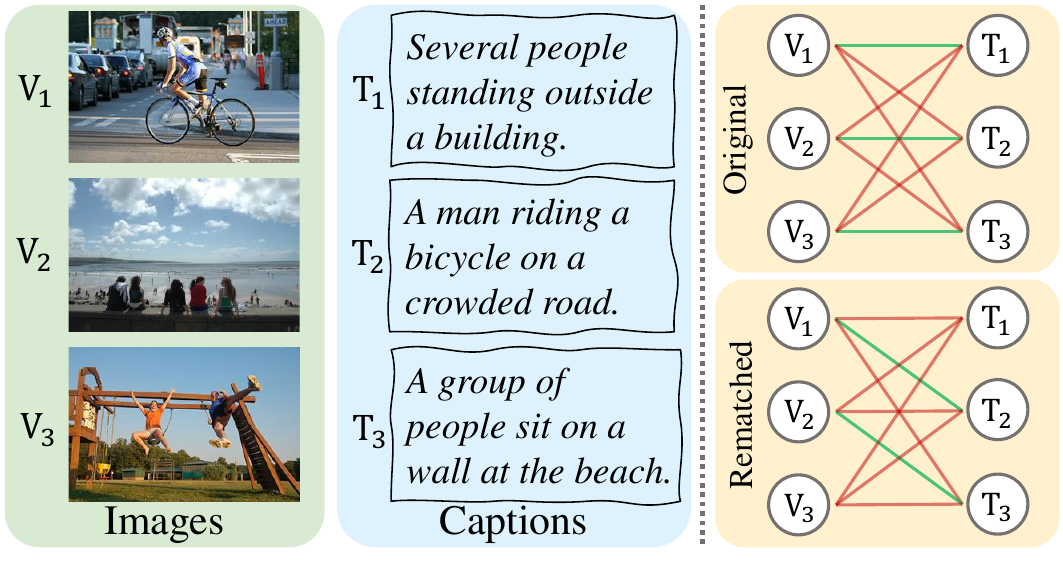}
\caption{A toy example to illustrate our idea. The potential semantic similarity among unpaired samples makes it possible to excavate useful knowledge from mismatched pairs. Our L2RM aims to rematch PMPs by generating a refined alignment that brings relevant cross-modal samples (green links) together while repelling irrelevant ones (red links) away from each other. We also show some real-world rematched cases for our L2RM in \cref{fig:case_study}.}
\label{fig:intro}
\vspace{-0.3cm} 
\end{figure}

To alleviate the PMP problem, existing works \cite{huang2021learning, han2023noisy, yang2023bicro} typically resort to recasting the estimated soft correspondence into a soft margin to adjust the distance in triplet ranking loss. However, the underuse of mismatched pairs, only limited to down-weighting their contribution, has led to sub-optimal retrieval performance. Hence, it is necessary to address the PMP issue in a data-efficient manner.

A question naturally arises: \emph{Could cross-modal retrieval models even learn useful knowledge from mismatched pairs?} To answer this question, this paper presents L2RM, a general framework that learns to rematch mismatched pairs for robust cross-modal retrieval. As illustrated in \cref{fig:intro}, our key idea is to excavate the potential matching relationship among mismatched cross-modal samples. Specifically, we first identify possibly mismatched pairs from training data by modeling the per-sample loss distribution. Then, we formalize the rematching idea as an OT problem to generate a new set of refined alignments for mismatched pairs in every minibatch. Notably, the cost function plays a paramount role when applying OT, which is typically designed as feature-driven distance \cite{gu2022keypoint,chang2022unified,dan2021learning}. However, the over-dependence on representations has led to a cycle of self-reinforcing errors---the existence of PMPs can generate corrupted representations--in turn, preventing the effective transport plan. To handle this problem, we propose a self-supervised learning solution to automatically learn the cost function from explicit similarity-cost mapping relation, which is unexplored in previous OT literature. Moreover, instead of exactly rematching all mismatched samples, we suggest modeling a partial OT problem while restricting the transport among false positives to boost the refined alignment. In practice, we show that our optimization objective could be solved by the Sinkhorn algorithm \cite{cuturi2013sinkhorn}, which only incurs cheap computational overheads.

Our main contributions are summarized as follows: (1) We propose a general OT-based framework to address the widely-existed PMP problem in cross-modal retrieval. The key to our method is learning to rematch mismatched pairs, which goes beyond previous efforts from the data-efficient view. (2) To address the error accumulation faced by the vanilla cost function, we propose a novel self-supervised learner that automatically learns the transport cost from explicit similarity-cost mapping relation. (3) To further boost the refined alignment, we present to
model a partial OT problem and restrict the transport among false positives. (4) Extensive experiments on several benchmarks demonstrate our L2RM endows existing cross-modal retrieval methods with strong robustness against PMPs.

\section{Related Work}
\label{sec:related work} 
\paragraph{Cross-Modal Retrieval.} 
Approaches for cross-modal retrieval aim to retrieve relevant items across different modalities for the query data. Current dominant methods project different modalities into a shared embedding space to measure the similarity of cross-modal pairs, which generally follow two research lines: 1) Global Alignment focuses on learning the correspondence between whole cross-modal data. Existing studies usually propose a two-stream network to learn comparable global features \cite{faghri2017vse++,zhang2020context,lu2022cots}. 2) Local Alignment. It seeks to align the fine-grained regions for more precise cross-modal matching. For example, \cite{li2019visual} employ the cross-attention mechanism to fully excavate the semantic region-word alignments. \cite{wei2020multi, xue2021probing} explore the intra-modal relation to facilitate inter-modal alignments.

Although these prior arts have achieved promising results, their success mainly relies on well-matched data, which is extremely expensive and even impossible to collect. To satisfy a more practical retrieval that is robust against the PMPs, \cite{huang2021learning, qin2022deep, yang2023bicro} divide the mismatched pairs from training data and estimate a soft correspondence to downweight their training contribution. Recently, \cite{hu2023cross} resorts to complementary contrastive learning that only utilizes the negative information to avoid overfitting. However, these methods neglect the usage of either the negative information \cite{huang2021learning, qin2022deep, yang2023bicro} or the positive one \cite{hu2023cross}. To fully leverage the training data, this paper proposes an OT-based method to rematch those partially mismatched pairs.

\paragraph{Optimal Transport.}
OT is used to seek a minimal-cost transport plan from one probability measure to another. The original OT model \cite{kantorovich2006translocation} is a linear program that incurs expensive computational cost. \cite{cuturi2013sinkhorn} proposes the entropy-regularized OT to provide a computationally cheaper solver. Recently, OT has gained increasing attention from different fields in machine learning, including unsupervised learning \cite{caron2020unsupervised}, semi-supervised learning \cite{tai2021sinkhorn}, object detection \cite{ge2021ota,afouras2022self}, domain adaptation \cite{redko2019optimal,fatras2022optimal}, and long-tailed recognition \cite{peng2022optimal,wang2022solar}. To the best of our knowledge, we are the first to perform the PMP problem from an OT perspective.

\section{Preliminaries}

\subsection{Background on OT}
OT provides a mechanism to infer the correspondence between two measures. We briefly introduce the OT theory to help us better view the PMP problem from an OT perspective. Consider $\bm{X} = \{{x}_{i} \}_{i=1}^{m}$ and $\bm{Y} = \{{y}_{j} \}_{j=1}^{n}$ as two discrete variables, and we denote their probability measures as $\bm{p}=\sum_{i=1}^{m}p_{i}\delta(x_{i})$ and $\bm{q}=\sum_{j=1}^{n}q_{j}\delta (y_{j})$, where $\delta$ is the Dirac function, $p_{i}$ and $q_{j}$ are the probability mass belonging to the probability simplex. When a meaningful cost function $c(\cdot)$ is defined, we can get the cost matrix $\bm{C} \in \mathbb{R}^{m\times n}$ between $\bm{X}$ and $\bm{Y}$, where $\bm{C}_{ij} = c(x_{i}, y_{j})$. Based on these, the OT distance can be expressed as:
\begin{equation}\label{eq:ot}
\begin{aligned}
&\ \text{OT}(\bm{p},\bm{q}) \triangleq \min_{\bm{\pi} \in \Pi(\bm{p},\bm{q})} \langle \bm{\pi} ,\bm{C} \rangle_F
\\ \ \mbox{ s.t. } \Pi(\bm{p},\bm{q}) &= \{\bm{\pi} \in \mathbb{R}^{m\times n}_+ \vert \bm{\pi} \mathbbm{1}_{n} = \bm{p}, \bm{\pi}^{\top} \mathbbm{1}_{m} = \bm{q}\},
\end{aligned}
\end{equation}
where $\langle \cdot,\cdot\rangle_F$ is the Frobenius dot-product and $\mathbbm{1}_{d}$ denotes a $d$-dimensional all-one vector. $\bm{\pi}$ is called the optimal transport plan that transport $\bm{p}$ towards $\bm{q}$ at the smallest cost.

\subsection{Problem Definition}
Without losing generality, we take the visual-text retrieval as an example to present the PMP problem in cross-modal retrieval. Consider a training dataset $\mathbbm{D}=\{(V_i, T_i,m_i) \}_{i=1}^N$ consisting of $N$ samples, where $(V_i, T_i)$ is the visual-text pair and $m_i \in \{1,0\}$ indicates whether the bimodal data is semantically matched or not. The key to cross-modal retrieval lies in measuring the similarity across distinct modalities. To achieve this, existing methods usually project the visual and textual modalities into a comparable feature space via the corresponding modal-specific networks $f_v$ and $f_t$, respectively. Then the similarity of a given visual-text pair is measured through $S_{ij} = g(f_v(V_i),f_t(T_j))$, where $g$ is a nonparametric or parametric mapping function. For convenience, we denote $g(f_v(V_i),f_t(T_j))$ as $g(V_i,T_j)$ in the following. 

Ideally, the positive (matched) pairs should have higher similarity while the negative (mismatched) pairs should have lower ones, which can be achieved by minimizing the triplet loss \cite{faghri2017vse++} or InfoNCE loss \cite{oord2018representation}. Consider a batch of $N_b$ pairs $\{(V_i,T_i)\}_{i=1}^{N_b}$, the triplet loss is defined as:
\begin{equation}\label{eq:triplet_loss}
\begin{aligned}
\mathcal{L}^{\text{triplet}}(V_i, T_i) =  &[\alpha - g(V_i,T_i) + g(V_i,\hat{T}_h)]_{+}
\\  + &[\alpha - g(V_i,T_i) + g(\hat{V}_h,T_i)]_{+},
\end{aligned}
\end{equation}
where $\alpha$ is a margin and $[x]_+ = max(x,0)$. $\hat{V}_h$ and $\hat{T}_h$ are the most similar negatives in the given batch corresponding to $(V_i, T_i)$. Eq.\eqref{eq:triplet_loss} aims to enforce the negative pairs to be distant from the positives by a certain margin value.

Alternatively, InfoNCE loss is extended to cross-modal scenario \cite{radford2021learning, hu2023cross} that encourages the similarity gap between positives and negatives as large as possible. Formally, the matching probability of $j$-th textual sample w.r.t. the $i$-th visual query is defined as $p_{ij}^{v2t} = \frac{\text{exp}(g(V_i,T_j)/\tau)}{\sum_{j^{\prime}=1}^{N_b}\text{exp}(g(V_i,T_{j^{\prime}})/\tau)}$, where $\tau$ is a temperature parameter. As InfoNCE loss is symmetric, the matching probability
$p_{ij}^{t2v}$ is defined similarly.
For notation convenience, we denote $\bm{p}_i^{v2t} = [p_{i1}^{v2t},\cdots,p_{iN_b}^{v2t}]^\top$ and $\bm{p}_i^{t2v} = [p_{i1}^{t2v},\cdots,p_{iN_b}^{t2v}]^\top$ as the probability vectors. To align cross-modal samples, the corresponding one-hot vector $\bm{y}_i = [y_{i1},\cdots,y_{iN_b}]^\top$ is used as supervision, where $y_{ij}$ equal to 1 if $i=j$ while other elements are 0. Thus, the cross-modal InfoNCE loss is given by:
\begin{equation}\label{eq:InfoNCE_loss}
\mathcal{L}^{\text{InfoNCE}}(V_i, T_i) = \mathcal{H}(\bm{y}_i,\bm{p}_i^{v2t}) + \mathcal{H}(\bm{y}_i,\bm{p}_i^{t2v}),
\end{equation}
where $\mathcal{H}$ is the batched cross-entropy function.

The success of both Eq.\eqref{eq:triplet_loss} and Eq.\eqref{eq:InfoNCE_loss} relies on the well-matched pairs. However, in practice, the multimedia datasets are usually web-collected, and thus inevitably contains an unknown portion of irrelevant pairs but are wrongly treated as matched $(m_i = 1)$. Our goal is to combat such PMPs to facilitate robust cross-modal retrieval.

\section{Methodology}
To tackle the PMP problem, the mainstream pipeline first uses the memorization effect \cite{arpit2017closer} of DNNs, \ie{}, DNNs learn simpler patterns before memorizing the difficult ones, to partition the dataset into a matched subset $\mathbbm{D}_m$, and a mismatched subset $\mathbbm{D}_{\widetilde m} = \mathbbm{D} / \mathbbm{D}_m$. After that, $\mathbbm{D}_m$ can be used for standard cross-modal training. To mitigate the impact of PMPs, recent advances \cite{huang2021learning, han2023noisy, yang2023bicro} introduce a soft margin into Eq.\eqref{eq:triplet_loss} to down-weight the samples from $\mathbbm{D}_{\widetilde m}$. However,  due to the underuse of mismatched pairs, the achieved performance by them is argued to be sub-optimal. In this work, we aim to fully leverage PMPs by trying to excavate the potential semantic similarity among mismatched pairs.  In the following, we present the details of our method.

\subsection{Identifying Mismatched Pairs}
Following the mainstream learning style,  we first identify possibly mismatched pairs from all training data. The memorization effect of DNNs indicates that mismatched samples tend to have relatively higher loss during the early stage of training. Based on this, we use the difference in loss distribution between the matched and mismatched pairs to divide the training set. Empirically, we observe that the distribution of triplet loss is more distinguishable. Thus, given the retrieval model $(f_v, f_t, g)$, we compute the per-sample loss through Eq.\eqref{eq:triplet_loss}:
\begin{equation}\label{eq:per_sample_loss}
\ell_{(f_v, f_t, g)} = \{\ell_i \}_{i=1}^N = \{\mathcal{L}^{\text{triplet}}(V_i, T_i)\}_{i=1}^N.
\end{equation}
Then, we fit a two-component beta mixture model \cite{arazo2019unsupervised, han2023noisy, yang2023bicro} to $\ell_{(f_v, f_t, g)}$ using the Expectation-Maximization algorithm. For $i$-th pair, its probability $w_i$ being mismatched is the posterior probability $p(b|\ell_i)$, where $b$ is the beta component with a higher mean. By setting a threshold on $\{ w_i\}_{i=1}^N$, we can divide the training data into the matched subset $\mathbbm{D}_m$ and mismatched subset $\mathbbm{D}_{\widetilde m}$ (we set the threshold to 0.5 in all experiments for brevity).

\begin{figure*}[!t]
  \centering
  \includegraphics[width=1\textwidth]{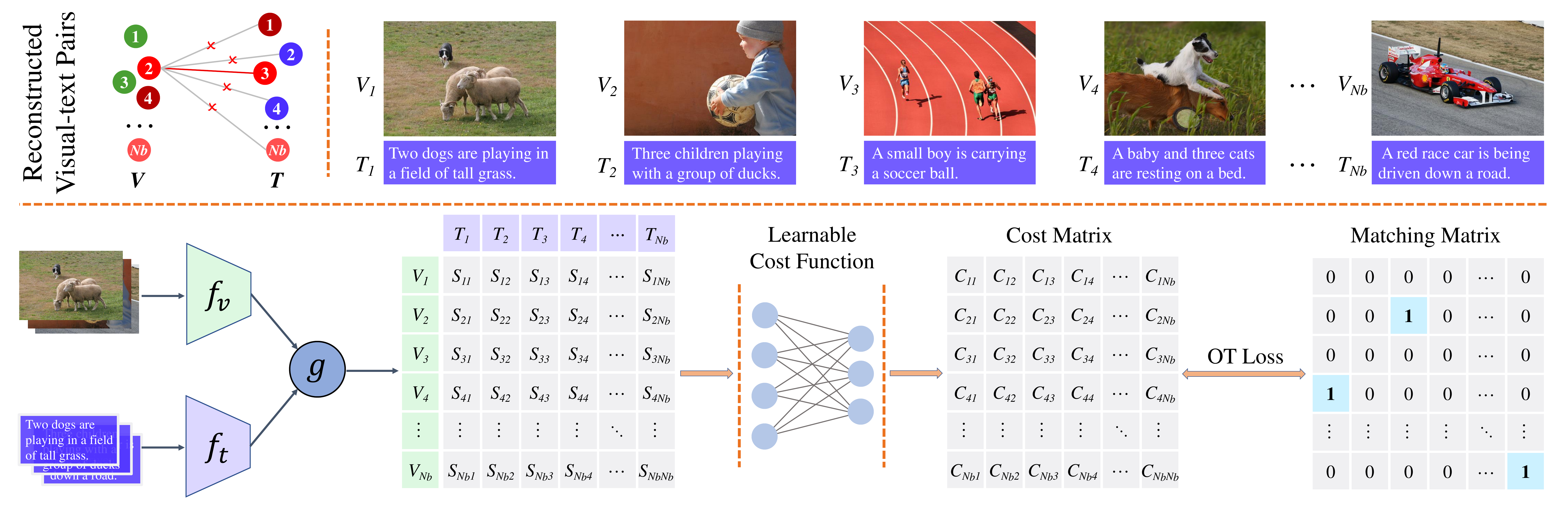}
  \caption{Overview of the learnable cost function with self-supervised learning. The up part illustrates the reconstructed pairs that only $(V_4,T_1)$, $(V_2,T_3)$, and $(V_{N_b},T_{N_b})$ are the reserved matching ones. Then, the matching matrix is viewed as supervision to guide the cost function from the explicit similarity-cost mapping relation through an OT loss (the down part).
  } \label{fig:cost_function}
\vspace{-0.1cm} 
\end{figure*}

For initial convergence of the algorithm, we warm up the model for a few epochs by training on all data with Eq.\eqref{eq:triplet_loss} or Eq.\eqref{eq:InfoNCE_loss}. However, for extreme mismatching rates, the model would quickly overfit to mismatched pairs and produce unreliable loss. To address this issue, we mitigate the overconfidence of the model by adding a reverse cross entropy \cite{wang2019symmetric} term to the InfoNCE loss during warm-up, \ie{},
\begin{equation}\label{eq:rce_loss}
\mathcal{L}^{\text{RCE}}(V_i, T_i) = \mathcal{H}(\bm{p}_i^{v2t},\bm{y}_i) + \mathcal{H}(\bm{p}_i^{t2v},\bm{y}_i).
\end{equation}
In the presence of PMPs, $\bm{y}_i$ may provide the wrong matching relation. Instead, the estimated probability could reflect the truer distribution to a certain extent. Note that we bound the one-hot label into $[\epsilon, 1-\epsilon]$ for computational feasibility.
($\epsilon = 10^{-7}$ in our experiments).

\subsection{Rematching Mismatched Pairs}
We formalize the rematching idea as an OT problem, generating refined alignments by seeking a minimal-cost transport plan. We will first introduce the novel learnable cost function to suit the PMP scenario, then we show how to boost the refined alignment by a relaxed OT model. 

\paragraph{Cost Function with Self-Supervised Learning.}
Cost function plays a crucial role when learning the transport plan for OT. In general, $\bm{C}_{ij}$ is set to a distance measure, \eg{}, $L_2$-distance \cite{gu2022keypoint} or cosine distance \cite{dan2021learning} to measure the expense of transporting a visual sample $i$ to a textual sample $j$. However, the existence of PMPs imposes formidable obstacles for these feature-driven distance measures. On the one hand, training with PMPs can wrongly bring irrelevant data together, which undoubtedly prevents effective representation learning. Even worse, different modalities will be embedded into separate regions of the shared space due to the inherent modality gap \cite{liang2022mind}.
On the other hand, the refined alignments produced by those corrupted features would be used to guide subsequent training, leading to the cycle of self-reinforcing errors \cite{chen2023two}. 

To address the aforementioned limitations, we propose a novel self-supervised learning approach to automatically learn the cost function. Intuitively, for a given image and caption, the transport cost can be modeled as a function of similarity that higher similarity enjoys a lower cost. Thus, we formulate the cost function as a single-layer feed-forward network with parameters $\Theta_c$, \ie{}, $f_c\left(;\Theta_c \right)$, which takes the similarity matrix of the batched visual-text samples as input and attempts to learn the corresponding cost matrix. To achieve this, we reconstruct the visual-text pairs to guide the cost function from explicit similarity-cost mapping relation. Specifically, for the matched pairs sampled from $\mathbbm{D}_m$, we randomly reserve a part of the matching images and substitute the images from $\mathbbm{D}_{\widetilde m}$ for the remaining ones. With the reserved indexes, we could automatically obtain a matching matrix that indicates the ideal matching probability for each reconstructed pair. For the example illustrated in \cref{fig:cost_function}, $(V_4,T_1)$, $(V_2,T_3)$, and $(V_{N_b},T_{N_b})$ are the reserved matching pairs with a matching probability of 1, while the others could be considered as mismatched ones with a matching probability of 0. For convenience, let $\mathbbm{D}^{\prime}$ be the reconstructed data, and $(\bm{V},\bm{T}) \in \mathbbm{D}^{\prime}$ be matrices that contain a batch of images and captions. To relate the similarity-cost mapping with the matching matrix, we optimize the cost function by the following OT loss:
\begin{equation}\label{eq:OT_loss}
\mathcal{L}_{\text OT}(\bm{\pi}^{\text{sup}}, \bm{V}, \bm{T}) = \langle \bm{\pi}^{\text{sup}} ,f_c\left( g\left( \bm{V},\bm{T} \right);\Theta_c \right)  \rangle_F ,
\end{equation}
where $\bm{\pi}^{\text{sup}}$ is the matching matrix, and $g\left( \bm{V},\bm{T} \right)$ denotes the similarity matrix for the batched visual-text pairs.

Eq.\eqref{eq:OT_loss} seeks an effective cost function from a reverse perspective of OT, which views the ideal transport plan as the supervision to minimize the transport cost.

\begin{figure*}[!t]
  \centering
  \includegraphics[width=1\textwidth]{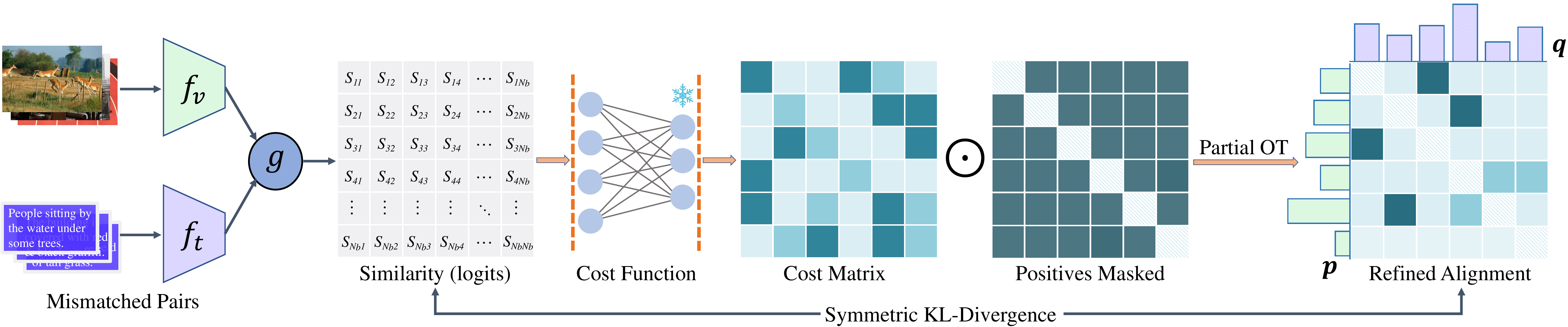}
  \caption{Illustration of the proposed rematching loss. For the mismatched pairs, we formalize a partial OT problem with positive pairs masked to generate the refined alignment in each batch. The refined alignment provides a more reliable matching relation to supervise the mismatched pairs. Then, we compute the symmetric KL-divergence to optimize the retrieval model $(f_v,f_t,g)$.
  } \label{fig:rematching}
\vspace{-0.4cm} 
\end{figure*}

\paragraph{Boosting Refined Alignments with Relaxed OT.}
Given the defined cost function, we could generate the refined alignments for mismatched pairs following the OT objective described in Eq.\eqref{eq:ot}. However, Eq.\eqref{eq:ot} requires the two distributions to have the same total mass and that all the mass of $\bm{p}$ should be transported to exactly match the mass of $\bm{q}$. In practice, due to the limited batch size, one caption may be irrelevant to all images in the batch and vice versa. To this end, we adopt the partial OT model \cite{figalli2010optimal, chapel2020partial} to relax such strict all-to-all mass constraints, which seeks a minimal cost of only transporting $0 \leqslant \rho \leqslant \min \left(\Vert \bm{p} \Vert_1, \Vert \bm{q} \Vert_1 \right)$ unit mass between the visual and textual distribution, \ie{},
\begin{align}\label{eq:pot}
\min_{\bm{\pi} \in \Pi^{\rho}(\bm{p},\bm{q})} \langle \bm{\pi} ,\bm{C} \rangle_F, &
\mbox{ s.t. } \Pi^{\rho} (\bm{p},\bm{q}) = \{\bm{\pi} \in \mathbb{R}^{m\times n}_+ \vert \bm{\pi} \mathbbm{1}_{n} \leqslant \bm{p}, \nonumber \\
\ &\bm{\pi}^{\top} \mathbbm{1}_{m} \leqslant \bm{q}, \mathbbm{1}_{m}^{\top}\pi \mathbbm{1}_{n} = \rho\}.
\end{align}

Furthermore, the false positives contained in the mismatched pairs introduce an implicit constraint to our transport plan $\bm{\pi}$ that the transport mass between the same element in two distributions should be limited. To this end, we propose to impose a mask operation on the transport plan that restricts the transport to only concentrate among the unpaired pairs. Specifically, the mask matrix $\bm{M} \in \mathbb{R}^{m\times n}$ is defined as:
\begin{equation}\label{eq:mask}
    M_{ij} \triangleq \begin{cases} 0, & \mbox{ if } i = j,\\
    1, & \text{otherwise}.\\
    \end{cases}
\end{equation}
Then the masked transport plan is defined as the Hadamard product $\tilde{\bm{\pi}} = \bm{M} \odot \bm{\pi}$ and be optimized through Eq.\eqref{eq:pot}.

\subsection{The Training Objective}

Given the mismatched pairs $\{(V_i, T_i)\}_{i=1}^{N_b}$ sampled from $\mathbbm{D}_{\widetilde m}$, if we don't have any prior knowledge, we could consider the visual and textual samples follow the uniform distributions, \ie{}, $\bm{p} = \sum_{i=1}^{N_b} \frac{1}{N_b}\delta(V_i)$ and $\bm{q} = \sum_{i=1}^{N_b} \frac{1}{N_b}\delta(T_i)$, respectively. To guarantee the efficiency of our algorithm, we adopt an online strategy to update $\Theta_c$ and calculate $\tilde{\bm{\pi}}$ through a single optimization loop:

\begin{equation}
\begin{aligned}\label{eq:bi-level}
    &\min\limits_{{\tilde{\bm{\pi}} \in \Pi^{\rho}(\bm{p}, \bm{q})}} \mathbb{E}_{(\bm{V},\bm{T})\in \mathbbm{D}_{\tilde m}} \left\langle \tilde{\bm{\pi}}, f_c \left( g\left( \bm{V},\bm{T} \right);\Theta_c^{*} \right)\right\rangle_{F} -\lambda H(\tilde{\bm{\pi}}),  \\
    &\ \ \ \ \ \ \mbox{ s.t. } \Theta_c^{*}=\argmin\limits_{\Theta_c} \mathbb{E}_{(\bm{V},\bm{T},\bm{\pi}^{\text {sup}})\in \mathbbm{D}^{\prime}} \mathcal{L}_{\text OT}(\bm{\pi}^{\text{sup}}, \bm{V}, \bm{T}) 
\end{aligned}
\end{equation}
where $\lambda > 0$ is a regularization parameter for the entropic constraint $H(\tilde{\bm{\pi}}) = -\sum_{ij} \tilde{\bm{\pi}}_{ij} \log \tilde{\bm{\pi}}_{ij}$. Note that Eq.\eqref{eq:bi-level} introduces an entropy regularization item to the OT model, which enables the transport plan to be solved by the computationally cheaper Sinkhorn-Knopp algorithm \cite{cuturi2013sinkhorn}. The detailed solution is presented in Appendix \red{A}.


The optimal transport plan from Eq.\eqref{eq:bi-level} represents a refined alignment that provides a more reliable matching relation for those mismatched visual-text samples. As our refined alignment is generated dynamically, we adopt the KL-divergence to compute the rematching loss instead of the cross entropy. Besides, a reverse term is added to symmetrize the KL-divergence, which makes the training more stable. Formally, let $\tilde{\bm{\pi}}_i^{v2t}$ and $\tilde{\bm{\pi}}_i^{t2v}$ be the row-wise and column-wise normalized refined alignment for the $i$-th sample, respectively. Then, the rematching loss (see \cref{fig:rematching}) is defined as:
\begin{equation}\label{eq:rematching-loss}
\begin{aligned}
\mathcal{L}^{\text{re}}\left( V_i, T_i\right) = &\frac{1}{2} \left[D_{K\!L}(\tilde{\bm{\pi}}_i^{v2t} \parallel \bm{p}_i^{v2t}) +  D_{K\!L}(\bm{p}_i^{v2t} \parallel \tilde{\bm{\pi}}_i^{v2t})\right] \\
+ &\frac{1}{2} \left[D_{K\!L}(\tilde{\bm{\pi}}_i^{t2v} \parallel \bm{p}_i^{t2v}) +  D_{K\!L}(\bm{p}_i^{t2v} \parallel \tilde{\bm{\pi}}_i^{t2v})\right].
\end{aligned}
\end{equation}

For the pairs that are divided as matched, we use the triplet ranking loss to directly control the distance gap. Thus, our final objective function is defined as:
\begin{equation}\label{eq:final}
\mathcal{L}^{\text{Final}} = \ \sum_{\mathclap{\left( V_i, T_i\right) \in \mathbbm{D}_m}}\  \mathcal{L}^{\text{triplet}}\left( V_i, T_i\right)  + \ \sum_{\mathclap{\left( V_i, T_i\right) \in \mathbbm{D}_{\tilde m}}}\  \mathcal{L}^{\text{re}}\left( V_i, T_i\right).
\end{equation}
The detailed training pseudo-code is shown in Appendix \red{B}.

\begin{table*}[htbp]
  \setlength {\belowcaptionskip} {-0.2cm}
  \centering
  \scalebox{1}{
    \small
    \addtolength{\tabcolsep}{-2.3pt}  
    \renewcommand{\arraystretch}{0.89}
    \begin{tabular}{c|l|ccc|ccc|c|ccc|ccc|c}
    \hline
    \hline
\multirow{3}{*}{MRate} & \multirow{3}{*}{Method} & \multicolumn{7}{c|}{Flickr30K}                        & \multicolumn{7}{c}{MS-COCO} \\
\cline{3-16}          &       & \multicolumn{3}{c|}{Image-to-Text} & \multicolumn{3}{c|}{Text-to-Image} & \multirow{2}{*}{rSum} & \multicolumn{3}{c|}{Image-to-Text} & \multicolumn{3}{c|}{Text-to-Image} & \multirow{2}{*}{rSum} \\
\cline{3-8}\cline{10-15}          &       & R@1   & R@5   & R@10  & R@1   & R@5   & R@10  &       & R@1   & R@5   & R@10  & R@1   & R@5   & R@10  &  \\
    \hline
    \multirow{10}{*}{0.2} 
          & IMRAM & 59.1  & 85.4  & 91.9  & 44.5  & 71.4  & 79.4  & 431.7 & 69.9  & 93.6  & 97.4  & 55.9  & 84.4  & 89.6  & 490.8 \\
          & NCR   & 73.5  & 93.2  & 96.6  & 56.9  & 82.4  & 88.5  & 491.1 & 76.6  & 95.6  & 98.2  & 60.8  & 88.8  & 95.0  & 515.0 \\
          & BiCro & 74.7  & 94.3  & 96.8  & 56.6  & 81.4  & 88.2  & 492.0 & 76.6  & 95.4  & 98.2  & 61.3  & 88.8  & 94.8  & 515.1 \\
          & DECL-SGR & 74.5  & 92.9  & 97.1  & 53.6  & 79.5  & 86.8  & 484.4 & 75.6  & 95.1  & 98.3  & 59.9  & 88.3  & 94.7  & 511.9 \\
          & DECL-SGRAF & 77.5  & 93.8  & 97.0  & 56.1  & 81.8  & 88.5  & 494.7 & 77.5  & 95.9  & 98.4  & 61.7  & 89.3  & 95.4  & 518.2 \\
          & RCL-SGR & 74.2  & 91.8  & 96.9  & 55.6  & 81.2  & 87.5  & 487.2 & 77.0  & 95.5  & 98.1  & 61.3  & 88.8  & 94.8  & 515.5 \\
          & RCL-SGRAF & 75.9  & 94.5  & 97.3  & 57.9  & 82.6  & 88.6  & 496.8 & 78.9 & 96.0  & 98.4  & 62.8  & 89.9  & 95.4  & 521.4 \\
          & L2RM-SAF & 73.7  & 94.3  & 97.7  & 56.8  & 81.8  & 88.1  & 492.4 & 77.9  & 96.0  & 98.3  & 62.1  & 89.2  & 94.9  & 518.4 \\
          & L2RM-SGR & 76.5  & 93.7  & 97.3  & 55.5  & 81.5  & 88.0  & 492.5 & 78.4  & 95.7  & 98.3  & 62.1  & 89.1  & 94.9  & 518.5 \\
          & L2RM-SGRAF & \textbf{77.9} & \textbf{95.2} & \textbf{97.8} & \textbf{59.8} & \textbf{83.6} & \textbf{89.5} & \textbf{503.8} & \textbf{80.2} & \textbf{96.3} & \textbf{98.5} & \textbf{64.2} & \textbf{90.1} & \textbf{95.4} & \textbf{524.7} \\
    \hline
    \multirow{10}{*}{0.4} 
          & IMRAM & 44.9  & 73.2  & 82.6  & 31.6  & 56.3  & 65.6  & 354.2 & 51.8  & 82.4  & 90.9  & 38.4  & 70.3  & 78.9  & 412.7 \\
          & NCR   & 68.1  & 89.6  & 94.8  & 51.4  & 78.4  & 84.8  & 467.1 & 74.7  & 94.6  & 98.0  & 59.6  & 88.1  & 94.7  & 509.7 \\
          & BiCro & 70.7  & 92.0  & 95.5  & 51.9  & 77.7  & 85.4  & 473.2 & 75.2  & 95.3  & 98.1  & 60.0  & 87.8  & 94.3  & 510.7 \\
          & DECL-SGR & 69.0  & 90.2  & 94.8  & 50.7  & 76.3  & 84.1  & 465.1 & 73.6  & 94.6  & 97.9  & 57.8  & 86.9  & 93.9  & 504.7 \\
          & DECL-SGRAF & 72.7  & 92.3  & 95.4  & 53.4  & 79.4  & 86.4  & 479.6 & 75.6  & 95.5  & 98.3  & 59.5  & 88.3  & 94.8  & 512.0 \\
          & RCL-SGR & 71.3  & 91.1  & 95.3  & 51.4  & 78.0  & 85.2  & 472.3 & 73.9  & 94.9  & 97.9  & 59.0  & 87.4  & 93.9  & 507.0 \\
          & RCL-SGRAF & 72.7  & 92.7  & 96.1  & 54.8  & 80.0  & 87.1  & 483.4 & 77.0  & 95.5  & 98.3  & 61.2  & 88.5  & 94.8  & 515.3 \\
          & L2RM-SAF & 72.1  & 92.1  & 96.1  & 52.7  & 78.8  & 85.9  & 477.7 & 74.4  & 94.7  & 98.3  & 59.2  & 87.9  & 94.4  & 508.9 \\
          & L2RM-SGR & 73.1  & 92.4  & 96.3  & 52.3  & 79.4  & 86.3  & 479.8 & 75.2  & 94.8  & 98.1  & 59.4  & 87.8  & 94.1  & 509.4 \\
          & L2RM-SGRAF & \textbf{75.8} & \textbf{93.2} & \textbf{96.9} & \textbf{56.3} & \textbf{81.0} & \textbf{87.3} & \textbf{490.5} & \textbf{77.5} & \textbf{95.8} & \textbf{98.4} & \textbf{62.0} & \textbf{89.1} & \textbf{94.9} & \textbf{517.7} \\
    \hline
    \multirow{10}{*}{0.6} 
          & IMRAM & 16.4  & 38.2  & 50.9  & 7.5   & 19.2  & 25.3  & 157.5 & 18.2  & 51.6  & 68.0  & 17.9  & 43.6  & 54.6  & 253.9 \\
          & NCR   & 13.9  & 37.7  & 50.5  & 11.0  & 30.1  & 41.4  & 184.6 & 0.1   & 0.3   & 0.4   & 0.1   & 0.5   & 1.0   & 2.4 \\
          & BiCro & 64.1  & 87.1  & 92.7  & 47.2  & 74.0  & 82.3  & 447.4 & 73.2  & 93.9  & 97.6  & 57.5  & 86.3  & 93.4  & 501.9 \\
          & DECL-SGR & 64.5  & 85.8  & 92.6  & 44.0  & 71.6  & 80.6  & 439.1 & 69.7  & 93.4  & 97.5  & 54.5  & 85.2  & 92.6  & 492.9 \\
          & DECL-SGRAF & 65.2  & 88.4  & 94.0  & 46.8  & 74.0  & 82.2  & 450.6 & 73.0  & 94.2  & 97.9  & 57.0  & 86.6  & 93.8  & 502.5 \\
          & RCL-SGR & 62.3  & 86.3  & 92.9  & 45.1  & 71.3  & 80.2  & 438.1 & 71.4  & 93.2  & 97.1  & 55.4  & 84.7  & 92.3  & 494.1 \\
          & RCL-SGRAF & 67.7  & 89.1  & 93.6  & 48.0  & 74.9  & 83.3  & 456.6 & 74.0  & 94.3  & 97.5  & 57.6  & 86.4  & 93.5  & 503.3 \\
          & L2RM-SAF & 66.1  & 88.8  & 93.8  & 47.8  & 74.2  & 82.2  & 452.9 & 71.2  & 93.4  & 97.5  & 56.5  & 85.9  & 93.0  & 497.5 \\
          & L2RM-SGR & 65.1  & 87.8  & 93.6  & 47.0  & 73.5  & 81.5  & 448.5 & 72.7  & 93.9  & 97.5  & 56.9  & 86.2  & 93.3  & 500.5 \\
          & L2RM-SGRAF & \textbf{70.0} & \textbf{90.8} & \textbf{95.4} & \textbf{51.3} & \textbf{76.4} & \textbf{83.7} & \textbf{467.6} & \textbf{75.4} & \textbf{94.7} & \textbf{97.9} & \textbf{59.2} & \textbf{87.4} & \textbf{93.8} & \textbf{508.4} \\
    \hline
    \multirow{10}{*}{0.8} 
          & IMRAM & 3.1   & 9.7   & 5.2   & 0.3   & 0.9   & 1.9   & 21.1  & 1.3   & 5.0   & 8.3   & 0.2   & 0.6   & 1.3   & 16.7 \\
          & NCR   & 1.5   & 6.2   & 9.9   & 0.3   & 1.0   & 2.1   & 21.0  & 0.1   & 0.3   & 0.4   & 0.1   & 0.5   & 1.0   & 2.4 \\
          & BiCro & 2.3   & 9.2   & 17.2  & 2.6   & 10.2  & 16.8  & 58.3  & 62.2  & 88.6  & 94.6  & 47.4  & 79.2  & 88.5  & 460.5 \\
          & DECL-SGR & 44.4  & 72.6  & 82.0  & 33.9  & 59.5  & 69.0  & 361.4 & 60.0  & 88.7  & 94.5  & 45.9  & 78.8  & 88.3  & 456.2 \\
          & DECL-SGRAF & 53.4  & 78.8  & 86.9  & 37.6  & 63.8  & 73.9  & 394.4 & 64.8  & 90.5  & 96.0  & 49.7  & 81.7  & 90.3  & 473.0 \\
          & RCL-SGR & 47.1  & 70.5  & 79.4  & 30.3  & 56.1  & 66.3  & 349.7 & 63.2  & 89.3  & 95.2  & 47.6  & 78.7  & 88.0  & 462.0 \\
          & RCL-SGRAF & 51.7  & 75.8  & 84.4  & 34.5  & 61.2  & 70.7  & 378.3 & 67.4  & 90.8  & 96.0  & 50.6  & 81.0  & 90.1  & 475.9 \\
          & L2RM-SAF & 50.8  & 77.9  & 85.5  & 35.6  & 62.6  & 72.7  & 385.1 & 64.7  & 90.8  & 95.8  & 50.0  & 80.9  & 89.4  & 471.6 \\
          & L2RM-SGR & 50.5  & 77.2  & 83.9  & 34.2  & 61.1  & 71.6  & 378.5 & 65.2  & 90.3  & 96.1  & 49.8  & 81.0  & 88.2  & 470.6 \\
          & L2RM-SGRAF & \textbf{55.7} & \textbf{80.8} & \textbf{87.8} & \textbf{39.4} & \textbf{65.4} & \textbf{74.9} & \textbf{404.0} & \textbf{69.0} & \textbf{91.9} & \textbf{96.4} & \textbf{52.6} & \textbf{82.4} & \textbf{90.3} & \textbf{482.6} \\
    \hline
    \hline
    \end{tabular}%
    }
  \caption{Image-text retrieval performance under different mismatching rates (MRate) on Flickr30K and MS-COCO.}
  \label{tab:synthesis}%
\end{table*}%

\section{Experiment}
\label{sec:experiment}
In this section, we experimentally analyze the effectiveness of L2RM in robust cross-modal retrieval.

\subsection{Setup}
\paragraph{Datasets.} 
We apply our method to three image-text retrieval datasets varying in scale and scope. Specifically, Flickr30K \cite{young2014image} consists of 31,000 images with five corresponding text annotations for each image from the Flickr website. Following \cite{huang2021learning}, we split 1,000 images for validation, 1,000 images for testing, and the rest for training. MS-COCO \cite{lin2014microsoft} is a large-scale cross-modal dataset, which collects 123,287 images with five sentences each. Following \cite{huang2021learning}, we use 5,000 images for validation, 5,000 images for testing, and the rest for training. Conceptual Captions \cite{sharma2018conceptual} is a web-crawled large-scale dataset containing 3.3M one-to-one images and captions. Following \cite{huang2021learning}, we use the subset, \ie{}, CC152K to conduct experiments, which has 150,000 images for training, 1,000 images for validation, and 1,000 images for testing.

\vspace{-0.1cm}

\paragraph{Implementation Details.}
As a general method, L2RM could be directly applied to almost all cross-modal retrieval methods to improve their robustness. Following \cite{qin2022deep, hu2023cross}, we apply L2RM to SGR, SAF, and SGRAF for a comprehensive comparison. We evaluate the retrieval performance with the Recall@K (R@K) metric. Following \cite{huang2021learning}, we save the best performance checkpoint on the validation set w.r.t. the sum of the evaluation scores and report its results on the testing set. We follow the same training setting as \cite{huang2021learning}, our specific parameters setting can be found in Appendix \red{C.1}.

\paragraph{Baselines.} 
We compare L2RM with eight state-of-the-art cross-modal retrieval methods, including four general methods (\ie{}, IMRAM \cite{chen2020imram}, SGR, SAF, and SGRAF \cite{diao2021similarity}) and four robust learning methods against the PMPs (\ie{}, NCR \cite{huang2021learning}, DECL \cite{qin2022deep}, BiCro \cite{yang2023bicro}, and RCL \cite{hu2023cross}). Note that the original BiCro combines four models, \ie{}, two co-trained SGR, and two co-trained SAF. For a fair comparison, we report the results of 2 co-trained SGR for BiCro like \cite{huang2021learning}.


\subsection{Main Results}
In this section, we conduct comparison experiments with different mismatching rates on three datasets to evaluate the performance of our L2RM. As Flickr30K and MS-COCO are well-established datasets, we carry out experiments by generating the synthesized false positive pairs, \ie{}, the mismatching rate (MRate) increases from $0.2$ to $0.8$ in intervals of $0.2$. Following \cite{qin2022deep, hu2023cross}, we randomly select a specific percentage of images and randomly permute all their corresponding captions, which is more challenging and practical than the setting in \cite{huang2021learning, yang2023bicro}. For the web-collected dataset CC152K, which naturally contains about 3\% $ \sim $  20 \% unknown mismatched pairs \cite{sharma2018conceptual}. Thus we directly conduct experiments on it to evaluate the performance with real PMPs.

\paragraph{Results on Synthesized PMPs.}
\cref{tab:synthesis} shows the experimental results on Flickr30K and MS-COCO. Note that for MS-COCO, the results are computed by averaging over 5 folds of 1K test images like \cite{qin2022deep, hu2023cross}. Due to space limitation, we omit the results of some general methods (SGR, SAF, and SGRAF), and the comparison on original datasets (0 MRate), which could be found in Appendix \red{C.2}. From the results, we can find that L2RM achieves the best results on all metrics than the other state-of-the-art methods, which shows the superior robustness of L2RM against PMPs. Moreover, when the mismatching rate is high, \eg{}, 0.6 and 0.8, the improvement of L2RM is more evident, proving that excavating mismatched pairs could effectively facilitate robust cross-modal retrieval.

\paragraph{Results on Real-World PMPs.}
We validate our method on the real-world dataset CC152K, which contains an unknown portion of mismatched pairs. As shown in \cref{tab:table_real}, our method considerably outperforms the best baseline in terms of sum in retrieval by 9.8\%. Notably, our L2RM-SGR surpasses all SGR variants by a clear margin, achieving as much as a 16.9\% (rSum) absolute improvement over the best variant. It is because the real-world rematched pairs are more likely to involve only local alignments, \eg{}, \cref{fig:case_study}(e)-\cref{fig:case_study}(f), while the SGR model itself is adept at capturing the relationship between local alignments.

\begin{table}[h]
  \setlength {\belowcaptionskip} {-0.2cm}
  \centering
    \renewcommand\arraystretch{1}
    \small
    \addtolength{\tabcolsep}{-3.6pt}  
    \renewcommand{\arraystretch}{0.92}
    \begin{tabular}{l|ccc|ccc|c}
    \hline
    \hline
    \multirow{2}{*}{Method} & \multicolumn{3}{c|}{Image-to-Text} & \multicolumn{3}{c|}{Text-to-Image} & \multirow{2}{*}{rSum} \\
    \cline{2-7}          & R@1   & R@5   & R@10  & R@1   & R@5   & R@10  &  \\
    \hline
    IMRAM & 27.8  & 52.4  & 60.9  & 29.2  & 51.5  & 61.2  & 283.0 \\
    SAF   & 32.5  & 59.5  & 70.0  & 32.5  & 60.7  & 68.7  & 323.9 \\
    SGR   & 14.5  & 35.5  & 48.9  & 13.7  & 36.1  & 47.9  & 196.6 \\
    NCR   & 39.5  & 64.5  & 73.5  & 40.3  & 64.6  & 73.2  & 355.6 \\
    BiCro & 39.7  & 64.6  & 72.6  & 39.2  & 65.0  & 74.1  & 355.2 \\
    DECL-SAF & 36.6  & 63.0  & 73.3  & 38.5  & 63.2  & 73.5  & 348.1 \\
    DECL-SGR & 36.2  & 63.6  & 73.2  & 37.1  & 63.6  & 73.7  & 347.4 \\
    DECL-SGRAF & 39.0  & 66.1  & 75.5  & 40.7  & 66.3  & 76.7  & 364.3 \\
    RCL-SAF & 37.5  & 63.0  & 71.4  & 37.8  & 62.4  & 72.4  & 344.5 \\
    RCL-SGR & 38.3  & 63.0  & 70.4  & 39.2  & 63.2  & 72.3  & 346.4 \\
    RCL-SGRAF & 41.7  & 66.0  & 73.6  & 41.6  & 66.4  & 75.1  & 364.4 \\
    L2RM-SAF & 37.3  & 62.7  & 71.7  & 38.8  & 65.7  & 74.8  & 351.0 \\
    L2RM-SGR & 39.5  & 66.2  & \textbf{76.0}  & 41.8  & 65.9  & 74.9  & 364.3 \\
    L2RM-SGRAF & \textbf{43.0} & \textbf{67.5} & 75.7 & \textbf{42.8} & \textbf{68.0} & \textbf{77.2} & \textbf{374.2} \\
    \hline
    \hline
    \end{tabular}%

  \caption{Image-text retrieval performance on CC152K.}
  \label{tab:table_real}%
\end{table}%
\vspace{-0.3cm} 

\subsection{Ablation Study}

\paragraph{Impact of Each Component.} To study the influence of specific components in our method, we carry out the ablation study on the Flickr30K with $0.6$ MRate. Specifically, we ablate the contributions of three key components of L2RM, \ie{}, partial OT, positives masked, and the learnable cost function (we use the cosine distance to measure the cost instead). Besides, we compare L2RM with different formulas of rematching loss: KL-divergence and InfoNCE. From \cref{tab:ablation}, we observe the following conclusions: 1) The full L2RM could achieve the best overall performance, showing that all three components are important to improve the robustness against PMPs. 2) Using the learnable cost function substantially outperforms the variant with the cosine distance cost (\eg{},$+11.9$ in terms of the rSum), which signifies the simple feature-driven cost is sub-optimal to the PMP situation. 3) Formulating different rematching loss could also achieve decent results, which verifies the ability of L2RM to provide effective matching relations.

\begin{table}[htbp]
\setlength {\belowcaptionskip} {-0.2cm}
  \centering
  \resizebox{1\columnwidth}{!}{
    \addtolength{\tabcolsep}{-4pt}  
    \renewcommand{\arraystretch}{0.95}
    \begin{tabular}{l|ccc|ccc}
    \hline
    \hline
    \multirow{2}{*}{Ablation} & \multicolumn{3}{c|}{Image-to-Text} & \multicolumn{3}{c}{Text-to-Image} \\
\cline{2-7}          & R@1   & R@5   & R@10  & R@1   & R@5   & R@10 \\
    \hline
    L2RM  & \textbf{65.1} & \textbf{87.8} & \textbf{93.6} & \textbf{47.0} & 73.5  & \textbf{81.5} \\
    L2RM w KL-divergence & 64.7  & 87.6  & 93.2  & 46.7  & \textbf{74.0} & 81.5 \\
    L2RM w InfoNCE Loss & 64.9  & 87.5  & 93.5  & 46.0  & 72.9  & 80.9 \\
    L2RM w/o Partial OT & 62.2  & 86.4  & 91.5  & 44.7  & 68.6  & 72.6 \\
    L2RM w/o Positives Masked & 64.9  & 87.6  & 92.7  & 46.4  & 73.2  & 81.1 \\
    L2RM w/o Cost Function & 61.3  & 85.6  & 91.4  & 44.9  & 72.4  & 81.0 \\
    \hline
    \hline
    \end{tabular}%
    }
  \caption{Ablation studies on Flickr30K with 0.6 MRate.}
  \label{tab:ablation}%
\end{table}%
\vspace{-0.4cm} 

\paragraph{Parameter Analysis.} 
We now investigate the effect of the parameter $\rho$ by plotting the recall scores with incremental $\rho$ on Flickr30K. The figure shows that the overall performance tends to decrease as $\rho$ increases. We further analyze how untransported pairs benefit the model training in Appendix \red{C.3}. Experimentally, we find that the refined alignments for untransported pairs can be equivalent to the label smoothing strategy \cite{szegedy2016rethinking}.

\begin{figure}[H]
\setlength {\belowcaptionskip} {-0.1cm}
\centering
\subfloat[Image to Text]{\includegraphics[width=0.48\columnwidth]{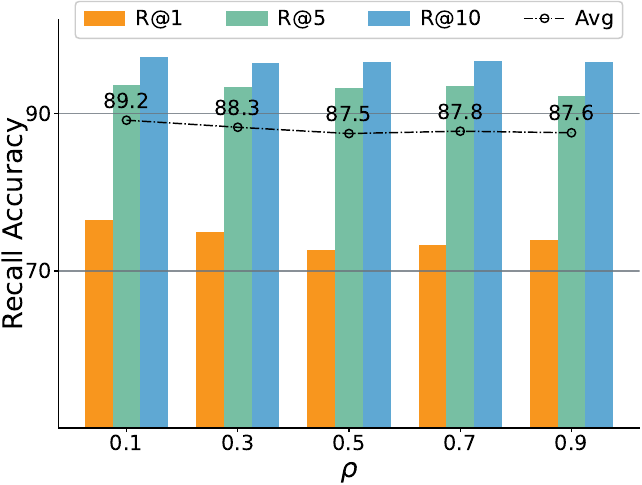}%
\label{pa_i2t}}
\subfloat[Text to Image]{\includegraphics[width=0.48\columnwidth]{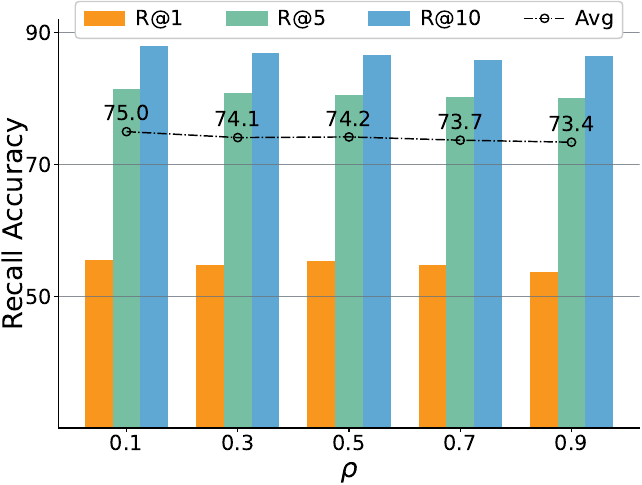}%
\label{pa_t2i}}
\caption{Parameter analysis of L2RM-SGR in terms of recall scores on the testing set of Flickr30K under 0.2 MRate.}
\label{fig:pa}
\end{figure}
\vspace{-0.2cm} 

\begin{table}[htbp]
\setlength {\belowcaptionskip} {-0.2cm}
  \centering
  \resizebox{1\columnwidth}{!}{
    \renewcommand\arraystretch{1.1}
    \addtolength{\tabcolsep}{-4pt}  
    \begin{tabular}{l|ccc|ccc|ccc|ccc}
    \hline
    \hline
    \multirow{3}{*}{Method} & \multicolumn{6}{c|}{Flickr30K}                & \multicolumn{6}{c}{CC152K} \\
\cline{2-13}          & \multicolumn{3}{c|}{Image-to-Text} & \multicolumn{3}{c|}{Text-to-Image} & \multicolumn{3}{c|}{Image-to-Text} & \multicolumn{3}{c}{Text-to-Image} \\
\cline{2-13}          & R@1   & R@5   & R@10  & R@1   & R@5   & R@10  & R@1   & R@5   & R@10  & R@1   & R@5   & R@10 \\
    \hline
    Triplet  & \large 0.1   & \large 0.7   & \large 1.3   & \large 0.1   & \large 0.7   & \large 1.1   & \large 38.2  & \large 64.2  & \large 71.4  & \large 39.5  & \large 64.2  & \large 71.9 \\
    InfoNCE & \large 45.9  & \large 71.4  & \large 80.3  & \large 29.2  & \large 52.7  & \large 61.1  & \large 39.3  & \large 66.8  & \large 75.0  & \large 40.8  & \large 65.2  & \large 74.2 \\
    \hline
    \hline
    \end{tabular}%
    }
  \caption{Comparison with different warm-up methods on Flickr30K with 0.8 MRate and CC152K.}
  \label{tab:warmup}%
\end{table}%
\vspace{-0.3cm} 

\begin{figure*}[t]
  \setlength {\belowcaptionskip} {-0.4cm}
  \centering
  \includegraphics[width=1\textwidth]{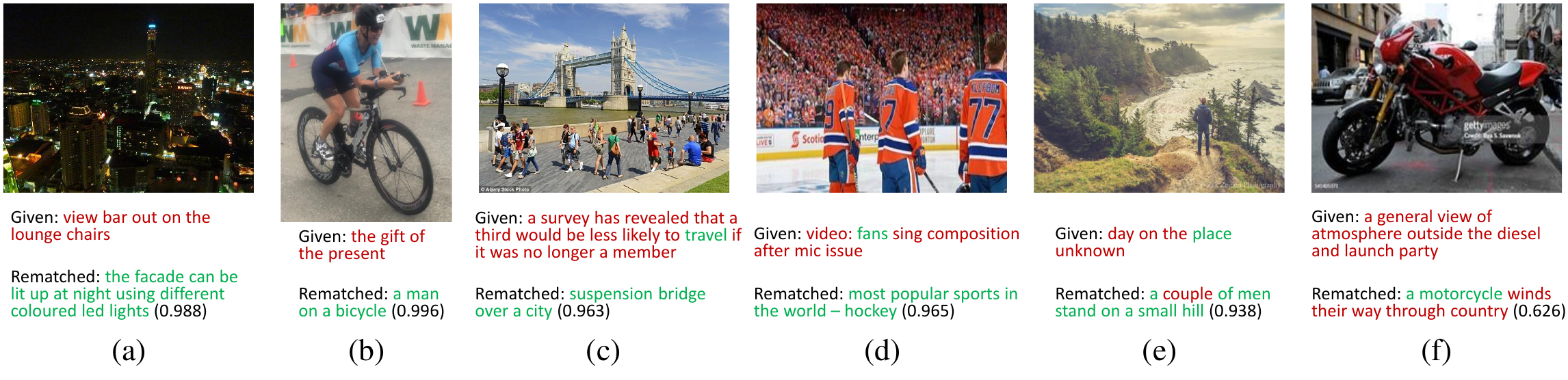}
  \caption{The ability of our L2RM to rematch the mismatched visual-text samples. The figure shows some representative rematched pairs for L2RM-SGR on the training set of CC152K dataset. We highlight the matched words in green and the mismatched words in red. } 
  \label{fig:case_study}
\end{figure*} 

\vspace{-0.2cm} 

\paragraph{Discussions on Warm-up Methods.} We use different warm-up methods, \ie{}, triplet loss \cite{faghri2017vse++} and InfoNCE loss \cite{oord2018representation} for our L2RM-SGR. The experiments are conducted on the Flickr30K with 0.8 MRate and the CC152K with real PMPs. As shown in \cref{tab:warmup}, one could see that the triplet loss cannot achieve satisfactory performance under the extreme mismatching rate. Compared with the results of the L2RM-SGR in \cref{tab:synthesis}, one could find that it is necessary to limit the overconfidence of the model during the warm-up process. The results on CC152K show that our method is robust to the choice of warm-up methods under a relatively low mismatching rate.

\subsection{Visualization and Analysis}

\begin{figure}[h]
\setlength {\belowcaptionskip} {-0.2cm}
\centering
\subfloat[Epoch 5]{\includegraphics[width=0.5\columnwidth]{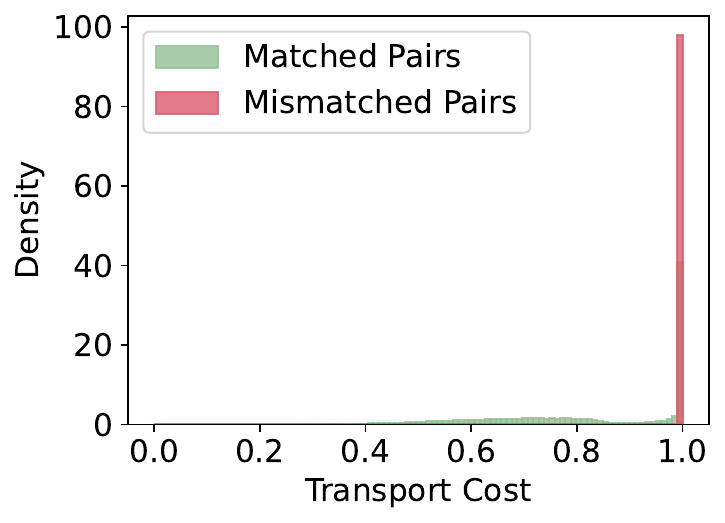}%
\label{epoch5}}
\subfloat[Epoch 35]{\includegraphics[width=0.5\columnwidth]{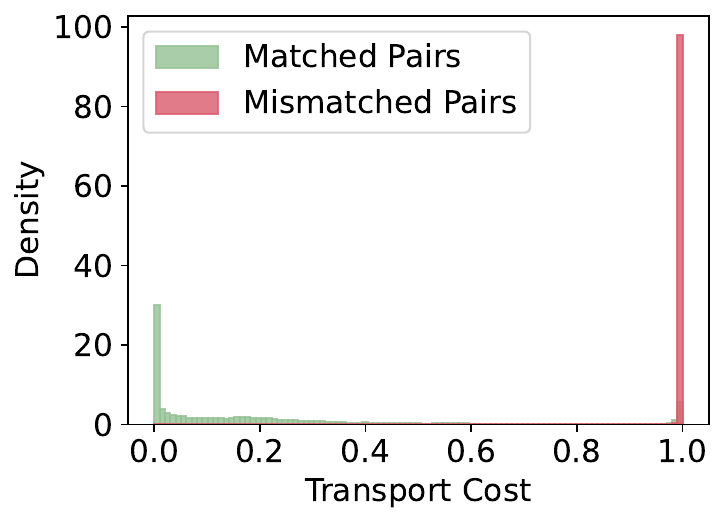}%
\label{epoch40}}
\caption{Transport cost distribution for matched and mismatched pairs at different training phases of our L2RM. The experiments are conducted on Flickr30K with 0.8 MRate.}
\label{fig:cost_visual}
\end{figure}

\paragraph{Distribution of Transport Cost.} 
To intuitively show the effectiveness of the learnable cost function, we illustrate the transport cost for matched and mismatched training pairs on Flickr30K with 0.8 MRate. From \cref{fig:cost_visual}, one could see that our cost function first learns to assign higher transport costs to those mismatched pairs. Although the costs of matched pairs are distributed over a large range in the early stage, they gradually become smaller and tend toward 0 as training proceeds. In conclusion, our cost function could successfully learn to distinguish matched and mismatched data, which lays the foundation for the further OT model.

\vspace{-0.4cm}

\paragraph{Visualizing Re-matched Image-Text Pairs.} 
To visually illustrate the rematching ability of our L2RM, we conduct the case study on CC152K to show the real-world rematched examples. Specifically, the first two rows of \cref{fig:case_study} show the image and its original mismatched caption, respectively. The third row shows the rematched caption provided by our method, and we also show the refined alignment scores in brackets. In particular, we could find that some real-world visual-text pairs are completely uncorrelated (\eg{}, \cref{fig:case_study}(a)-\cref{fig:case_study}(b)) or contain only a few local similarities (\eg{}, \cref{fig:case_study}(c)-\cref{fig:case_study}(e)). Thanks to our L2RM, the potential matching relation among mismatched pairs could be fully excavated to provide refined alignments. For example, one could see that the rematched caption, \ie{}, \textit{"a man on a bicycle"} nicely expresses the semantic concept in \cref{fig:case_study}(a). Although some rematched captions could not perfectly share the same semantics with images, they also contain some local similarities to the given images. For example, the image in \cref{fig:case_study}(f) is correctly described with the words \textit{"a motorcycle"} and our L2RM provides a relatively low refined alignment score as the target. In summary, our proposed rematching strategy could embrace better data efficiency and robustness against PMPs.


\section{Conclusion}
This work studies the challenge of cross-modal retrieval with partially mismatched pairs (PMPs). To address this problem, we propose L2RM, a generalized OT-based framework that learns to rematch mismatched pairs. Our key idea is to excavate the potential semantic similarity among unpaired samples. To formalize this idea through OT, first, we propose a self-supervised learner to automatically learn effective cost function. Second, we model a partial OT problem and restrict the transport among false positives to further boost refined alignments. Extensive experiments are conducted to verify that our L2RM can endow cross-modal retrieval models with strong robustness against PMPs.



{\small
\bibliographystyle{ieee_fullname}
\bibliography{egbib}
}
\appendix
\appendixtitle

\DeclareRobustCommand*{\IEEEauthorrefmark}[1]{%
    \raisebox{0pt}[0pt][0pt]{\textsuperscript{\footnotesize\ensuremath{#1}}}}

\def\thesection{\Alph{section}}



\LinesNumberedHidden
\begin{algorithm}[t]
\begin{minipage}{0.99\textwidth}
\caption{The training pipeline of our L2RM.}
\label{algo_pipeline}

\KwIn{The training dataset $\mathbbm{D}$ with PMPs, cross-modal retrieval model $(f_v,f_t,g)$, self-supervised learning cost function $f_c$, partial transport parameter $\rho$, Sinkhorn regularization parameter $\lambda$.}

Warm up the model $(f_v,f_t,g)$ using $\mathcal{L}^{\text{InfoNCE}} + \mathcal{L}^{\text{RCE}}$

\For{$e=1:num\_epochs$}
{
\tcp{identifying mismatched pairs}

$\mathcal{W} = \{ w_i\}_{i=1}^N \leftarrow BetaMixtureModel \left(\mathbbm{D},\left(f_v,f_t,g \right)\right)$

$\mathbbm{D}_m = \{ (V_i, T_i) \mid w_i \leq 0.5, \forall (V_i, T_i) \in \mathbbm{D} \}$,\ \  $\mathbbm{D}_{\widetilde m} = \{ (V_i, T_i) \mid w_i > 0.5, \forall (V_i, T_i) \in \mathbbm{D} \}$

\For{$n=1:num\_steps$}
{
\tcp{update the learnable cost function}
Reconstruct the visual-text pairs $\mathbbm{D}^{\prime}$ 

Sample a batched samples and get the corresponding matching matrix $(\bm{V}, \bm{T}, \bm{\pi}^{\text{sup}})$

Train the cost function $f_c$ on $(\bm{V}, \bm{T}, \bm{\pi}^{\text{sup}})$ by minimizing $\mathcal{L}_{\text OT}$

\tcp{rematching mismatched pairs}
Sample a batched samples $\mathbbm{B}_{\widetilde m} = \{ (V_i, T_i) \}_{i=1}^{N_b}$ from the mismatched subset $\mathbbm{D}_{\widetilde m}$

Compute the refined alignment $\tilde{\bm{\pi}}$ in the batch by optimizing the partial OT problem

\tcp{update the cross-modal retrieval model}
Sample a batched samples $\mathbbm{B}_{m} = \{ (V_i, T_i) \}_{i=1}^{N_b}$ from the matched subset $\mathbbm{D}_{m}$

Train the retrieval model $(f_v,f_t,g)$ on $(\mathbbm{B}_{m}, \mathbbm{B}_{\widetilde m})$ by minimizing $\mathcal{L}^{\text{Final}}$

}

}

\KwOut{Retrieval model $(f_v,f_t,g)$.}
\end{minipage}

\end{algorithm}

\section{Limitations}
Our work still has certain limitations, including (1) This work only explores the PMP problem among visual and textual modalities. Further research is needed to confirm the applicability of L2RM in other cross-modal domains against PMPs, \eg, re-identification \cite{qin2023noisy} and graph matching \cite{lin2023graph}. (2) The effectiveness of our rematched method is limited by the batch size. When using smaller batch sizes, the likelihood of observing semantic relevant pairs will decrease. One possible improvement is to maintain a queue to compare more data. We also provide experimental analysis (see \red{D.2} for details) to show the impact of batch size.

\section{Fast Solver for Refined Alignment}
In this section, we detail the fast approximation for computing the refined alignment. We will first introduce how to transform the original partial OT problem into a standard OT problem. Then, we will describe the solution by adopting the efficient Sinkhorn-Knopp algorithm. 

\paragraph{Transform partial OT to OT-like problem.} Recall that our partial OT problem seeks only $\rho$-unit mass of $\bm{p}=\sum_{i=1}^{m}p_{i}\delta(x_{i})$ and $\bm{q}=\sum_{j=1}^{n}q_{j}\delta (y_{j})$ is matched. To solve the exact partial OT problem, Chapel \etal{} \cite{chapel2020partial} propose an ingenious method that transforms the original partial OT problem into an OT-like problem. Specifically, consider two 
\vspace*{255 pt}

\noindent virtual samples $x_{m+1}$ and $y_{n+1}$ are added to the original variables $\bm{X}$ and $\bm{Y}$, respectively. Intuitively, to ensure $\rho$-unit mass is transported between $\{{x}_{i} \}_{i=1}^{m}$ and $\{{y}_{j} \}_{j=1}^{n}$, we should constrain the transport mass from $\{{x}_{i} \}_{i=1}^{m}$ to $y_{n+1}$ to $\lVert \bm{p} \rVert_1 - \rho$ and the transport mass from $\{{y}_{j} \}_{j=1}^{n}$ to $x_{m+1}$ to $\lVert \bm{q} \rVert_1 - \rho$. Thus, the original partial OT problem from $\bm{X} = \{{x}_{i} \}_{i=1}^{m}$ to $\bm{Y} = \{{y}_{j} \}_{j=1}^{n}$ can be transformed into a standard OT problem from $\hat{\bm{X}} = \{{x}_{i} \}_{i=1}^{m+1}$ to $\hat{\bm{Y}} = \{{y}_{j} \}_{j=1}^{n+1}$, where the corresponding probability measures are extended to $\hat{\bm{p}} = [ \bm{p}^\top, \lVert \bm{q} \rVert_1 - \rho]^\top$ and $\hat{\bm{q}} = [ \bm{q}^\top, \lVert \bm{p} \rVert_1 - \rho]^\top$, respectively. Following \cite{chapel2020partial}, the original cost matrix $\bm{C}$ is extended to $\hat{\bm{C}} \in \mathbb{R}^{m+1\times n+1}$:
\begin{equation}
\hat{\bm{C}} =
\begin{bmatrix}
\bm{C} & \xi \mathbbm{1}_{n} \\
\xi \mathbbm{1}_{m}^{\top} & 2\xi + A
\end{bmatrix},
\end{equation}
where $A>\max (\bm{C}_{ij})$ and $\xi >0$. Note that our original partial OT problem restricts the transport among the false positive pairs by imposing a mask matrix, which is extended by:
\begin{equation}
\hat{\bm{M}} =
\begin{bmatrix}
\bm{M} &  \mathbbm{1}_{n} \\
\mathbbm{1}_{m}^{\top} & 1
\end{bmatrix}.
\end{equation}
Based on these, computing the optimal transport plan in partial OT boils down to solve the following problem:
\begin{equation}\label{eq:ot_like}
\begin{aligned}
& \ \ \ \ \ \ \ \ \ \ \ \ \ \ \ \ \ \ \ \ \ \ \min_{\hat{\bm{\pi}} \in \Pi(\hat{\bm{p}},\hat{\bm{q}};\hat{\bm{M}})} \langle \hat{\bm{M}} \odot \hat{\bm{\pi}} ,\hat{\bm{C}} \rangle_F
\\ & \mbox{ s.t. } \Pi(\hat{\bm{p}},\hat{\bm{q}};\hat{\bm{M}}) = \{\hat{\bm{\pi}} \in \mathbb{R}^{m+1\times n+1}_+ \vert (\hat{\bm{M}} \odot \hat{\bm{\pi}}) \mathbbm{1}_{n} = \hat{\bm{p}},
\\ & \ \ \ \ \ \ \ \ \ \ \ \ \ \ \ \ \ \ \ \ \ \ \ \ \ \ \ \ \ \ \ \ (\hat{\bm{M}} \odot \hat{\bm{\pi}})^{\top} \mathbbm{1}_{m} = \hat{\bm{q}}\}.
\end{aligned}
\end{equation}
Eq.\eqref{eq:ot_like} is a standard OT problem and our objective $\tilde{\bm{\pi}} = (\hat{\bm{M}} \odot \hat{\bm{\pi}})[1:m, 1:n]$.

\paragraph{Solving OT with Sinkhorn algorithm.} Exactly solving the OT problem with linear programming algorithms requires high computational overhead. To resolve Eq.\eqref{eq:ot_like} efficiently, we resort to the entropy-regularized OT problem by adding a entropic constraint $-\lambda H(\hat{\bm{M}} \odot \hat{\bm{\pi}})$, which enables the transport plan to be computed by the lightspeed Sinkhorn-Knopp algorithm \cite{cuturi2013sinkhorn}. Note that Gu \etal{} \cite{gu2022keypoint} show that the Sinkhorn's algorithm can be applied to solve the transport plan with mask operation. The detailed solution is presented in Algorithm.~\ref{algo_SK}. We can see that the Sinkhorn's iteration only contains matrix multiplication and exponential operations, which can be computed efficiently.

\LinesNumberedHidden
\begin{algorithm}[t]

\caption{Solving Eq.(\red{3}) with Sinkhorn algorithm.}
\label{algo_SK}

\KwIn{Distribution $\hat{\bm{p}}$ and $\hat{\bm{q}}$, cost matrix $\hat{\bm{C}}$, mask matrix $\hat{\bm{M}}$, partial transport mass $\rho$, Sinkhorn regularization parameter $\lambda$, max iterations $it_{max}$.}

Initialize $\hat{\bm{K}} = \hat{\bm{M}} \odot e^{\frac{-\hat{\bm{C}}}{\lambda}}$, $\bm{b} \gets \mathbbm{1}_{n+1}, it \gets 0$

\tcp{Run Sinkhorn iterations}
\While{$it \leq  it_{max}$ and $\bm{a}$, $\bm{b}$ not convergence}{

\vspace{0.5em}

$\bm{a} \gets \frac{\hat{\bm{p}}}{\hat{\bm{K}} \bm{b}}$ \tcp{element-wise division}

\vspace{0.5em}
$\bm{b} \gets \frac{\hat{\bm{q}}}{\hat{\bm{K}}^{\top}\bm{a}} $
}
\tcp{Get the approximate solution}
$\hat{\bm{\pi}} = \text{diag}(\bm{a}) \hat{\bm{K}} \text{diag}(\bm{b}) $

\KwOut{Refined alignment  $\tilde{\bm{\pi}} = (\hat{\bm{M}} \odot \hat{\bm{\pi}})[1:m, 1:n]$.}

\end{algorithm}


\section{Training Pipeline}
In this section, we summarize our detailed training pipeline in Algorithm.~\ref{algo_pipeline}. The code of L2RM is available at \url{https://github.com/hhc1997/L2RM}.

\section{Additional Experiments}
\subsection{Implementation Details}
\paragraph{Input preprocessing.} Our experiments used the same input preprocessing as in the evaluation of NCR \cite{huang2021learning}. Specifically, all raw images are processed into the top 36 region proposals by the Faster-RCNN, where each is encoded as a 2048-dimensional feature.

\paragraph{Backbone architecture.} L2RM is a general framework which could endow almost all existing cross-modal retrieval methods robust against PMPs. Same as previous robust methods \cite{huang2021learning, qin2022deep, yang2023bicro, hu2023cross}, we implement L2RM based on SGR, SAF, and SGRAF \cite{diao2021similarity}. Specifically, the image regions and captions are projected into a common representation space by a full-connected network (\ie{},$f_v$) and a Bi-GRU model ((\ie{},$f_t$)), respectively. To calculate the cross-modal similarities, the similarity function $g$ is based on the Similarity Graph Reasoning (SGR), Similarity Attention Filtration (SAF), or the combination of SGR and SAF.

\begin{table}[htbp]
\setlength{\abovecaptionskip}{0.2cm}  
\setlength {\belowcaptionskip} {-0.1cm}
  \centering
  
  \resizebox{.9\columnwidth}{!}{
    \renewcommand\arraystretch{.9}
    \small
    \addtolength{\tabcolsep}{-4.5pt}  
    \begin{tabular}{l|c|c|c}
    \hline
    \hline
    Epochs & Flickr30K & MS-COCO & CC152K \\
    \hline
    warm up & 5     & 10    & 10 \\
    training  & 35    & 20    & 40 \\
    total & 40    & 30    & 50 \\
    update learning rate & 15    & 10    & 20 \\
    \hline
    \hline
    \end{tabular}%
    }
  \caption{The epoch settings for training on three datasets.}
  \label{tab:epoch}%
\end{table}%
\vspace{-0.1cm}

\begin{table*}[htbp]
  \setlength {\belowcaptionskip} {-0.3cm}
  \centering
  \scalebox{1}{
    \small
    \addtolength{\tabcolsep}{-1.5pt}  
    \renewcommand{\arraystretch}{0.9}
    \resizebox{0.98\textwidth}{!}{%
    \begin{tabular}{c|l|ccc|ccc|c|ccc|ccc|c}
    \multicolumn{1}{r}{} & \multicolumn{1}{r}{} &       &       & \multicolumn{1}{r}{} &       &       & \multicolumn{1}{r}{} & \multicolumn{1}{r}{} &       &       & \multicolumn{1}{r}{} &       &       & \multicolumn{1}{r}{} &  \\
    \hline
    \hline
    \multirow{3}{*}{MRate} & \multirow{3}{*}{Method} & \multicolumn{7}{c|}{Flickr30K}                        & \multicolumn{7}{c}{MS-COCO} \\
\cline{3-16}          &       & \multicolumn{3}{c|}{Image-to-Text} & \multicolumn{3}{c|}{Text-to-Image} & \multirow{2}{*}{rSum} & \multicolumn{3}{c|}{Image-to-Text} & \multicolumn{3}{c|}{Text-to-Image} & \multirow{2}{*}{rSum} \\
\cline{3-8}\cline{10-15}          &       & R@1   & R@5   & R@10  & R@1   & R@5   & R@10  &       & R@1   & R@5   & R@10  & R@1   & R@5   & R@10  &  \\
    \hline
        \multirow{14}{*}{0.2} & IMRAM & 59.1  & 85.4  & 91.9  & 44.5  & 71.4  & 79.4  & 431.7 & 69.9  & 93.6  & 97.4  & 55.9  & 84.4  & 89.6  & 490.8 \\
          & SAF   & 62.8  & 88.7  & 93.9  & 49.7  & 73.6  & 78.0  & 446.7 & 71.5  & 94.0  & 97.5  & 57.8  & 86.4  & 91.9  & 499.1 \\
          & SGR   & 55.9  & 81.5  & 88.9  & 40.2  & 66.8  & 75.3  & 408.6 & 25.7  & 58.8  & 75.1  & 23.5  & 58.9  & 75.1  & 317.1 \\
          & NCR   & 73.5  & 93.2  & 96.6  & 56.9  & 82.4  & 88.5  & 491.1 & 76.6  & 95.6  & 98.2  & 60.8  & 88.8  & 95.0  & 515.0 \\
          & BiCro & 74.7  & 94.3  & 96.8  & 56.6  & 81.4  & 88.2  & 492.0 & 76.6  & 95.4  & 98.2  & 61.3  & 88.8  & 94.8  & 515.1 \\
          & DECL-SAF & 73.4  & 92.0  & 96.4  & 53.6  & 79.7  & 86.4  & 481.5 & 74.4  & 95.3  & 98.2  & 59.8  & 88.3  & 94.8  & 510.8 \\
          & DECL-SGR & 74.5  & 92.9  & 97.1  & 53.6  & 79.5  & 86.8  & 484.4 & 75.6  & 95.1  & 98.3  & 59.9  & 88.3  & 94.7  & 511.9 \\
          & DECL-SGRAF & 77.5  & 93.8  & 97.0  & 56.1  & 81.8  & 88.5  & 494.7 & 77.5  & 95.9  & 98.4  & 61.7  & 89.3  & 95.4  & 518.2 \\
          & RCL-SAF & 72.0  & 91.7  & 95.8  & 53.6  & 79.9  & 86.7  & 479.7 & 77.1  & 95.5  & 98.2  & 61.0  & 88.8  & 94.6  & 515.2 \\
          & RCL-SGR & 74.2  & 91.8  & 96.9  & 55.6  & 81.2  & 87.5  & 487.2 & 77.0  & 95.5  & 98.1  & 61.3  & 88.8  & 94.8  & 515.5 \\
          & RCL-SGRAF & 75.9  & 94.5  & 97.3  & 57.9  & 82.6  & 88.6  & 496.8 & 78.9  & 96.0  & 98.4  & 62.8  & 89.9  & 95.4  & 521.4 \\
          & L2RM-SAF & 73.7  & 94.3  & 97.7  & 56.8  & 81.8  & 88.1  & 492.4 & 77.9  & 96.0  & 98.3  & 62.1  & 89.2  & 94.9  & 518.4 \\
          & L2RM-SGR & 76.5  & 93.7  & 97.3  & 55.5  & 81.5  & 88.0  & 492.5 & 78.4  & 95.7  & 98.3  & 62.1  & 89.1  & 94.9  & 518.5 \\
          & L2RM-SGRAF & \textbf{77.9} & \textbf{95.2} & \textbf{97.8} & \textbf{59.8} & \textbf{83.6} & \textbf{89.5} & \textbf{503.8} & \textbf{80.2} & \textbf{96.3} & \textbf{98.5} & \textbf{64.2} & \textbf{90.1} & \textbf{95.4} & \textbf{524.7} \\
    \hline
    \multirow{14}{*}{0.4} & IMRAM & 44.9  & 73.2  & 82.6  & 31.6  & 56.3  & 65.6  & 354.2 & 51.8  & 82.4  & 90.9  & 38.4  & 70.3  & 78.9  & 412.7 \\
          & SAF   & 7.4   & 19.6  & 26.7  & 4.4   & 12.0  & 17.0  & 87.1  & 13.5  & 43.8  & 48.2  & 16.0  & 39.0  & 50.8  & 211.3 \\
          & SGR   & 4.1   & 16.6  & 24.1  & 4.1   & 13.2  & 19.7  & 81.8  & 1.3   & 3.7   & 6.3   & 0.5   & 2.5   & 4.1   & 18.4 \\
          & NCR   & 68.1  & 89.6  & 94.8  & 51.4  & 78.4  & 84.8  & 467.1 & 74.7  & 94.6  & 98.0  & 59.6  & 88.1  & 94.7  & 509.7 \\
          & BiCro & 70.7  & 92.0  & 95.5  & 51.9  & 77.7  & 85.4  & 473.2 & 75.2  & 95.3  & 98.1  & 60.0  & 87.8  & 94.3  & 510.7 \\
          & DECL-SAF & 70.1  & 90.6  & 94.4  & 49.7  & 76.6  & 84.1  & 465.5 & 73.3  & 94.6  & 98.1  & 57.9  & 87.2  & 94.1  & 505.2 \\
          & DECL-SGR & 69.0  & 90.2  & 94.8  & 50.7  & 76.3  & 84.1  & 465.1 & 73.6  & 94.6  & 97.9  & 57.8  & 86.9  & 93.9  & 504.7 \\
          & DECL-SGRAF & 72.7  & 92.3  & 95.4  & 53.4  & 79.4  & 86.4  & 479.6 & 75.6  & 95.5  & 98.3  & 59.5  & 88.3  & 94.8  & 512.0 \\
          & RCL-SAF & 68.8  & 89.8  & 95.0  & 51.0  & 76.7  & 84.8  & 466.1 & 74.8  & 94.8  & 97.8  & 59.0  & 87.1  & 93.9  & 507.4 \\
          & RCL-SGR & 71.3  & 91.1  & 95.3  & 51.4  & 78.0  & 85.2  & 472.3 & 73.9  & 94.9  & 97.9  & 59.0  & 87.4  & 93.9  & 507.0 \\
          & RCL-SGRAF & 72.7  & 92.7  & 96.1  & 54.8  & 80.0  & 87.1  & 483.4 & 77.0  & 95.5  & 98.3  & 61.2  & 88.5  & 94.8  & 515.3 \\
          & L2RM-SAF & 72.1  & 92.1  & 96.1  & 52.7  & 78.8  & 85.9  & 477.7 & 74.4  & 94.7  & 98.3  & 59.2  & 87.9  & 94.4  & 508.9 \\
          & L2RM-SGR & 73.1  & 92.4  & 96.3  & 52.3  & 79.4  & 86.3  & 479.8 & 75.2  & 94.8  & 98.1  & 59.4  & 87.8  & 94.1  & 509.4 \\
          & L2RM-SGRAF & \textbf{75.8} & \textbf{93.2} & \textbf{96.9} & \textbf{56.3} & \textbf{81.0} & \textbf{87.3} & \textbf{490.5} & \textbf{77.5} & \textbf{95.8} & \textbf{98.4} & \textbf{62.0} & \textbf{89.1} & \textbf{94.9} & \textbf{517.7} \\
    \hline
    \multirow{14}{*}{0.6} & IMRAM & 16.4  & 38.2  & 50.9  & 7.5   & 19.2  & 25.3  & 157.5 & 18.2  & 51.6  & 68.0  & 17.9  & 43.6  & 54.6  & 253.9 \\
          & SAF   & 0.1   & 1.5   & 2.8   & 0.4   & 1.2   & 2.3   & 8.3   & 0.1   & 0.5   & 0.7   & 0.8   & 3.5   & 6.3   & 11.9 \\
          & SGR   & 1.5   & 6.6   & 9.6   & 0.3   & 2.3   & 4.2   & 24.5  & 0.1   & 0.6   & 1.0   & 0.1   & 0.5   & 1.1   & 3.4 \\
          & NCR   & 13.9  & 37.7  & 50.5  & 11.0  & 30.1  & 41.4  & 184.6 & 0.1   & 0.3   & 0.4   & 0.1   & 0.5   & 1.0   & 2.4 \\
          & BiCro & 64.1  & 87.1  & 92.7  & 47.2  & 74.0  & 82.3  & 447.4 & 73.2  & 93.9  & 97.6  & 57.5  & 86.3  & 93.4  & 501.9 \\
          & DECL-SAF & 56.6  & 82.5  & 89.7  & 40.4  & 66.6  & 76.6  & 412.4 & 68.6  & 92.9  & 97.4  & 54.1  & 84.9  & 92.7  & 490.6 \\
          & DECL-SGR & 64.5  & 85.8  & 92.6  & 44.0  & 71.6  & 80.6  & 439.1 & 69.7  & 93.4  & 97.5  & 54.5  & 85.2  & 92.6  & 492.9 \\
          & DECL-SGRAF & 65.2  & 88.4  & 94.0  & 46.8  & 74.0  & 82.2  & 450.6 & 73.0  & 94.2  & 97.9  & 57.0  & 86.6  & 93.8  & 502.5 \\
          & RCL-SAF & 63.9  & 84.8  & 91.7  & 43.0  & 71.2  & 79.4  & 434.0 & 70.1  & 93.1  & 96.8  & 54.5  & 84.4  & 91.9  & 490.8 \\
          & RCL-SGR & 62.3  & 86.3  & 92.9  & 45.1  & 71.3  & 80.2  & 438.1 & 71.4  & 93.2  & 97.1  & 55.4  & 84.7  & 92.3  & 494.1 \\
          & RCL-SGRAF & 67.7  & 89.1  & 93.6  & 48.0  & 74.9  & 83.3  & 456.6 & 74.0  & 94.3  & 97.5  & 57.6  & 86.4  & 93.5  & 503.3 \\
          & L2RM-SAF & 66.1  & 88.8  & 93.8  & 47.8  & 74.2  & 82.2  & 452.9 & 71.2  & 93.4  & 97.5  & 56.5  & 85.9  & 93.0  & 497.5 \\
          & L2RM-SGR & 65.1  & 87.8  & 93.6  & 47.0  & 73.5  & 81.5  & 448.5 & 72.7  & 93.9  & 97.5  & 56.9  & 86.2  & 93.3  & 500.5 \\
          & L2RM-SGRAF & \textbf{70.0} & \textbf{90.8} & \textbf{95.4} & \textbf{51.3} & \textbf{76.4} & \textbf{83.7} & \textbf{467.6} & \textbf{75.4} & \textbf{94.7} & \textbf{97.9} & \textbf{59.2} & \textbf{87.4} & \textbf{93.8} & \textbf{508.4} \\
    \hline
    \multirow{14}{*}{0.8} & IMRAM & 3.1   & 9.7   & 5.2   & 0.3   & 0.9   & 1.9   & 21.1  & 1.3   & 5.0   & 8.3   & 0.2   & 0.6   & 1.3   & 16.7 \\
          & SAF   & 0.0   & 0.8   & 1.2   & 0.1   & 0.5   & 1.1   & 3.7   & 0.2   & 0.8   & 1.4   & 0.1   & 0.5   & 1.0   & 4.0 \\
          & SGR   & 0.2   & 0.3   & 0.5   & 0.1   & 0.6   & 1.0   & 2.7   & 0.2   & 0.6   & 1.0   & 0.1   & 0.5   & 1.0   & 3.4 \\
          & NCR   & 1.5   & 6.2   & 9.9   & 0.3   & 1.0   & 2.1   & 21.0  & 0.1   & 0.3   & 0.4   & 0.1   & 0.5   & 1.0   & 2.4 \\
          & BiCro & 2.3   & 9.2   & 17.2  & 2.6   & 10.2  & 16.8  & 58.3  & 62.2  & 88.6  & 94.6  & 47.4  & 79.2  & 88.5  & 460.5 \\
          & DECL-SAF & 46.9  & 73.7  & 83.0  & 32.1  & 59.0  & 69.4  & 364.1 & 59.3  & 87.9  & 94.8  & 46.3  & 79.1  & 88.9  & 456.3 \\
          & DECL-SGR & 44.4  & 72.6  & 82.0  & 33.9  & 59.5  & 69.0  & 361.4 & 60.0  & 88.7  & 94.5  & 45.9  & 78.8  & 88.3  & 456.2 \\
          & DECL-SGRAF & 53.4  & 78.8  & 86.9  & 37.6  & 63.8  & 73.9  & 394.4 & 64.8  & 90.5  & 96.0  & 49.7  & 81.7  & 90.3  & 473.0 \\
          & RCL-SAF & 45.0  & 72.8  & 80.8  & 30.7  & 56.5  & 67.3  & 353.1 & 62.9  & 89.3  & 94.9  & 47.1  & 77.9  & 87.4  & 459.5 \\
          & RCL-SGR & 47.1  & 70.5  & 79.4  & 30.3  & 56.1  & 66.3  & 349.7 & 63.2  & 89.3  & 95.2  & 47.6  & 78.7  & 88.0  & 462.0 \\
          & RCL-SGRAF & 51.7  & 75.8  & 84.4  & 34.5  & 61.2  & 70.7  & 378.3 & 67.4  & 90.8  & 96.0  & 50.6  & 81.0  & 90.1  & 475.9 \\
          & L2RM-SAF & 50.8  & 77.9  & 85.5  & 35.6  & 62.6  & 72.7  & 385.1 & 64.7  & 90.8  & 95.8  & 50.0  & 80.9  & 89.4  & 471.6 \\
          & L2RM-SGR & 50.5  & 77.2  & 83.9  & 34.2  & 61.1  & 71.6  & 378.5 & 65.2  & 90.3  & 96.1  & 49.8  & 81.0  & 88.2  & 470.6 \\
          & L2RM-SGRAF & \textbf{55.7} & \textbf{80.8} & \textbf{87.8} & \textbf{39.4} & \textbf{65.4} & \textbf{74.9} & \textbf{404.0} & \textbf{69.0} & \textbf{91.9} & \textbf{96.4} & \textbf{52.6} & \textbf{82.4} & \textbf{90.3} & \textbf{482.6} \\
    \bottomrule
    \bottomrule
    \end{tabular}%
    }
   }
  \caption{Image-text retrieval performance under different mismatching rates (MRate) on Flickr30K and MS-COCO.}
  \label{tab:full_synthesis}%
\end{table*}%

\paragraph{Hyperparameters.} We follow the same training setting as NCR where applicable. Specifically, the word embedding size is 300 and the common space size is 1024. The retrieval model is trained by a Adam optimizer (default settings) with a learning rate of $2 \times 10^{-4}$ and a batch size of 128. The epoch setting for training is shown in \cref{tab:epoch}. The learning rate will be decayed by 0.1 when the training achieves the update epoch. The margin $\alpha$ used in triplet loss is fixed as 0.2 for all experiments.

For hyperparameters specific to L2RM, we set the temperature parameter $\tau$ as 0.05. We train our learnable cost function using the Adam optimizer with the default settings and a learning rate of $2 \times 10^{-6}$. To solve the OT problem, we fix the partial transport mass $\rho = 0.1$ for all experiments. Note that for the experiments conducted on original datasets (0 MRate), we empirically find that disabling the positives masked strategy could achieve superior performance. In addition, we set the Sinkhorn regularization parameter $\lambda$ as 0.01, 0.07, and 0.07 for Flickr30K, MS-COCO, and CC152K, respectively.

\subsection{More Comparisons Results}
\paragraph{Results under Synthesized PMPs.}
\cref{tab:full_synthesis} shows the full comparison results on Flickr30K and MS-COCO under different mismatching rates. From the results, one could see that the existence of PMPs remarkably impair the performance of general cross-modal retrieval methods (\ie{}, IMRAM, SAF, and SGR). With the mismatching rates increasing, their retrieval performance will degrade fast. Compared with the robust methods, we can find that our L2RM consistently outperforms them under different variants.

\begin{table*}[ht]
  \setlength {\belowcaptionskip} {-0.0cm}
  \centering
    \addtolength{\tabcolsep}{-2.5pt}  
    \renewcommand{\arraystretch}{.9}
    \begin{tabular}{l|ccc|ccc|c|ccc|ccc|c}
    \hline
    \hline
    \multirow{3}{*}{Method} & \multicolumn{7}{c|}{Flickr30K}                        & \multicolumn{7}{c}{MS-COCO} \\
\cline{2-15}          & \multicolumn{3}{c|}{Image-to-Text} & \multicolumn{3}{c|}{Text-to-Image} & \multirow{2}{*}{rSum} & \multicolumn{3}{c|}{Image-to-Text} & \multicolumn{3}{c|}{Text-to-Image} & \multirow{2}{*}{rSum} \\
\cline{2-7}\cline{9-14}          & R@1   & R@5   & R@10  & R@1   & R@5   & R@10  &       & R@1   & R@5   & R@10  & R@1   & R@5   & R@10  &  \\
    \hline
    IMRAM & 68.8  & 91.6  & 96.0  & 53.0  & 79.0  & 87.1  & 475.5 & 74.0  & 95.6  & 98.4  & 60.6  & 88.9  & 94.6  & 512.1 \\
    SAF   & 73.7  & 93.3  & 96.3  & 56.1  & 81.5  & 88.0  & 488.9 & 76.1  & 95.4  & 98.3  & 61.8  & 89.4  & 95.3  & 516.3 \\
    SGR   & 75.2  & 93.3  & 96.6  & 56.2  & 81.0  & 86.5  & 488.8 & 78.0  & 95.8  & 98.2  & 61.4  & 89.3  & 95.4  & 518.1 \\
    SGRAF & 77.8  & 94.1  & 97.4  & 58.5  & 83.0  & 88.8  & 499.6 & 79.6  & 96.2  & 98.5  & 63.2  & 90.7  & 96.1  & 524.3 \\
    NCR   & 77.3  & 94.0  & 97.5  & 59.6  & 84.4  & 89.9  & 502.7 & 78.7  & 95.8  & 98.5  & 63.3  & 90.4  & 95.8  & 522.5 \\
    BiCro & 79.5  & 94.2  & 97.4  & 59.4  & 83.6  & 89.8  & 503.9 & 78.4  & 95.6  & 98.5  & 62.6  & 89.7  & 95.7  & 520.5 \\
    DECL-SGRAF & 78.9  & 94.7  & 97.4  & 59.3  & 84.1  & 89.8  & 504.2 & 79.3  & 96.5  & 98.7  & 63.3  & 90.6  & 95.0  & 523.4 \\
    RCL-SAF & 76.7  & 93.7  & 97.3  & 56.2  & 82.6  & 88.8  & 495.3 & 78.5  & 96.1  & 98.6  & 62.7  & 90.0  & 95.4  & 521.3 \\
    RCL-SGR & 77.5  & 94.7  & 97.4  & 58.8  & 83.3  & 88.9  & 500.6 & 78.2  & 96.2  & 98.4  & 62.9  & 90.0  & 95.7  & 521.4 \\
    RCL-SGRAF & \textbf{79.9} & \textbf{96.1} & 97.8  & \textbf{61.1} & \textbf{85.4} & \textbf{90.3} & \textbf{510.6} & 80.4  & 96.4  & 98.7  & 64.3  & 90.8  & 96.0  & 526.6 \\
    L2RM-SAF & 77.1  & 93.2  & 96.7  & 57.5  & 82.4  & 87.8  & 494.7 & 78.2  & 95.7  & 98.6  & 63.4  & 89.6  & 95.1  & 520.6 \\
    L2RM-SGR & 79.1  & 94.1  & 97.7  & 58.1  & 83.6  & 88.9  & 501.5 & 79.0  & 96.4  & 98.3  & 63.7  & 90.2  & 95.8  & 523.4 \\
    L2RM-SGRAF & 79.6  & 95.9  & \textbf{98.4} & 60.7  & 84.8  & 89.0  & 508.4 & \textbf{80.5} & \textbf{96.6} & \textbf{98.9} & \textbf{65.7} & \textbf{90.8} & \textbf{96.1} & \textbf{528.6} \\
    \hline
    \hline
    \end{tabular}%

  \caption{Image-text retrieval performance on original Flickr30K and MS-COCO datasets.}
  \label{tab:table_orginal}%
\end{table*}%

\paragraph{Results on well-annotated Datasets.}
The Flickr30K and MS-COCO are two well-annotated datasets (almost 0 MRate), thus we conduct comparison experiments on the original Flickr30K and MS-COCO to show L2RM's performance under well-matched pairs. The experimental results are reported in \cref{tab:table_orginal}. From the results, one could observe that L2RM can boost the retrieval performance of existing methods, \ie{}, SAF, SGR, and SGRAF, even though it is proposed to improve robustness. On the one hand, the dataset cannot be absolutely well-matched; it still contains a few mismatched pairs. On the other hand, our rematching strategy augments more positive pairs to a certain extent by comparing unpaired samples, which could enhance the generalization of the model.

\paragraph{Results on MS-COCO 5K Datasets.}
\cref{tab:table_MSCOCO_5K_f} shows the quantitative results on MS-COCO with full 5K test images. From the results, we could observe that co-trained models offer bigger gains when the test data becomes complex.

\begin{table}[h]
  \centering
    \footnotesize
    \addtolength{\tabcolsep}{-3.9pt}  
    \renewcommand{\arraystretch}{1}
    \begin{tabular}{c|l|ccc|ccc|c}
    \hline
    \multirow{2}{*}{MRate} & \multirow{2}{*}{Method} & \multicolumn{3}{c|}{Image-to-Text} & \multicolumn{3}{c|}{Text-to-Image} & \multirow{2}{*}{rSum} \\
\cline{3-8}          &       & R@1   & R@5   & R@10  & R@1   & R@5   & R@10  &  \\
    \hline
    \multirow{3}[2]{*}{0.2} & L2RM-SAF & 56.6  & 83.3  & 90.9  & 40.1  & 69.5  & 80.0  & 420.4 \\
          & L2RM-SGR & 56.6  & 83.4  & 90.6  & 40.6  & 69.5  & 80.0  & 420.7 \\
          & L2RM-SGRAF & 59.6  & 85.1  & 92.0  & 42.5  & 71.5  & 81.3  & 432.0 \\
    \hline
    \multirow{3}[2]{*}{0.4} & L2RM-SAF & 53.1  & 81.6  & 89.8  & 38.4  & 67.5  & 78.2  & 408.6 \\
          & L2RM-SGR & 53.5  & 81.0  & 89.5  & 38.0  & 66.9  & 77.7  & 406.6 \\
          & L2RM-SGRAF & 57.1  & 83.4  & 91.0  & 40.8  & 69.4  & 79.7  & 421.4 \\
    \hline
    \multirow{3}[2]{*}{0.6} & L2RM-SAF & 51.0  & 78.4  & 86.8  & 34.9  & 63.1  & 74.7  & 388.9 \\
          & L2RM-SGR & 50.2  & 79.0  & 87.8  & 34.5  & 63.0  & 74.6  & 389.1 \\
          & L2RM-SGRAF & 53.5  & 81.0  & 88.9  & 37.3  & 65.7  & 76.7  & 403.1 \\
    \hline
    \multirow{3}[2]{*}{0.8} & L2RM-SAF & 40.7  & 71.2  & 80.9  & 28.2  & 55.8  & 68.0  & 344.8 \\
          & L2RM-SGR & 42.6  & 71.5  & 81.7  & 28.8  & 55.7  & 67.3  & 347.6 \\
          & L2RM-SGRAF & 45.7  & 74.4  & 83.9  & 30.9  & 58.5  & 69.8  & 363.2 \\
    \hline
    \end{tabular}%
  \caption{Performance under different MRates on MS-COCO 5K.}
  \label{tab:table_MSCOCO_5K_f}
\end{table}%

\paragraph{Impact of Batch Size.} To study the influence of different batch sizes for our method, we conducted the ablation study on Flickr30K with 0.6 MRate. Note that our method can flexibly adapt to different batch sizes by adjusting the transport mass $\rho$, and we set $\rho$ to 0.05, 0.1, and 0.2 for the batch size 64, 128, and 256, respectively. From \cref{tab:table_bs_f}, one could observe that our method still achieves superior results with a small batch size, \ie, 64, and even surpasses the second-best baseline RCL-SGRAF (in terms of the rSum metric) using a 128 batch size. We could also see that our L2RM can gain from a larger batch size, \ie, 256, while some methods may suffer a performance drop.

\begin{table}[t]
  \centering
    \small
    \footnotesize
    \addtolength{\tabcolsep}{-3.9pt}  
    \renewcommand{\arraystretch}{1}
    \begin{tabular}{c|l|ccc|ccc|c}
    \hline
    \multirow{2}{*}{Batch} & \multirow{2}{*}{Method} & \multicolumn{3}{c|}{Image-to-Text} & \multicolumn{3}{c|}{Text-to-Image} & \multirow{2}{*}{rSum} \\
\cline{3-8}          &       & R@1   & R@5   & R@10  & R@1   & R@5   & R@10  &  \\
    \hline
    \multirow{4}[1]{*}{64} & L2RM-SAF & 63.5  & 86.4  & 93.2  & 45.8  & 73.0  & 81.4  & 443.3 \\
          & L2RM-SGR & 62.9  & 87.4  & 92.7  & 46.1  & 72.8  & 81.3  & 443.2 \\
          & RCL-SGRAF & 66.9  & 88.3  & 94.1  & 48.3  & \textbf{75.3}  & 82.5  & 455.4 \\
          & L2RM-SGRAF & \textbf{67.2}  & \textbf{89.4}  & \textbf{94.2}  & \textbf{49.2}  & \textbf{75.3}  & \textbf{83.4}  & \textbf{458.7} \\
    \hline
    \multirow{4}[2]{*}{128} & L2RM-SAF & 66.1  & 88.8  & 93.8  & 47.8  & 74.2  & 82.2  & 452.9 \\
          & L2RM-SGR & 65.1  & 87.8  & 93.6  & 47.0  & 73.5  & 81.5  & 448.5 \\
          & RCL-SGRAF & 67.7  & 89.1  & 93.6  & 48.0  & 74.9  & 83.3  & 456.6 \\
          & L2RM-SGRAF & \textbf{70.0}  & \textbf{90.8}  & \textbf{95.4}  & \textbf{51.3}  & \textbf{76.4}  & \textbf{83.7}  & \textbf{467.6} \\
    \hline
    \multirow{4}[2]{*}{256} & L2RM-SAF & 66.7  & 89.0  & 93.5  & 48.0  & 74.2  & 82.1  & 453.5 \\
          & L2RM-SGR & 66.0  & 88.5  & 94.2  & 48.2  & 73.9  & 82.2  & 453.0 \\
          & RCL-SGRAF & 66.4  & 88.9  & 94.0  & 47.0  & 73.3  & 81.3  & 450.9 \\
          & L2RM-SGRAF & \textbf{69.7}  & \textbf{91.4}  & \textbf{95.6}  & \textbf{51.6}  & \textbf{77.1}  & \textbf{83.6}  & \textbf{469.0} \\
    \hline
    \end{tabular}%
  \caption{Performance with different batch sizes on Flickr30K.}
  \label{tab:table_bs_f}
\end{table}%

\begin{figure*}[ht]
	\centering
	\includegraphics[width=1\textwidth]{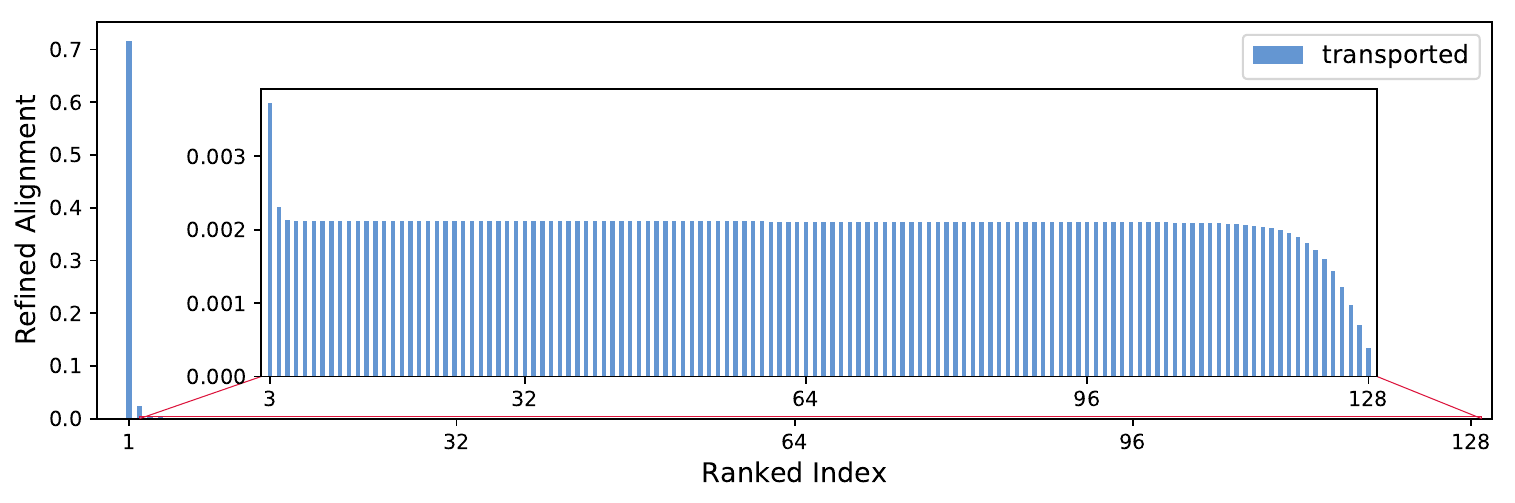} \\
	\includegraphics[width=1\textwidth]{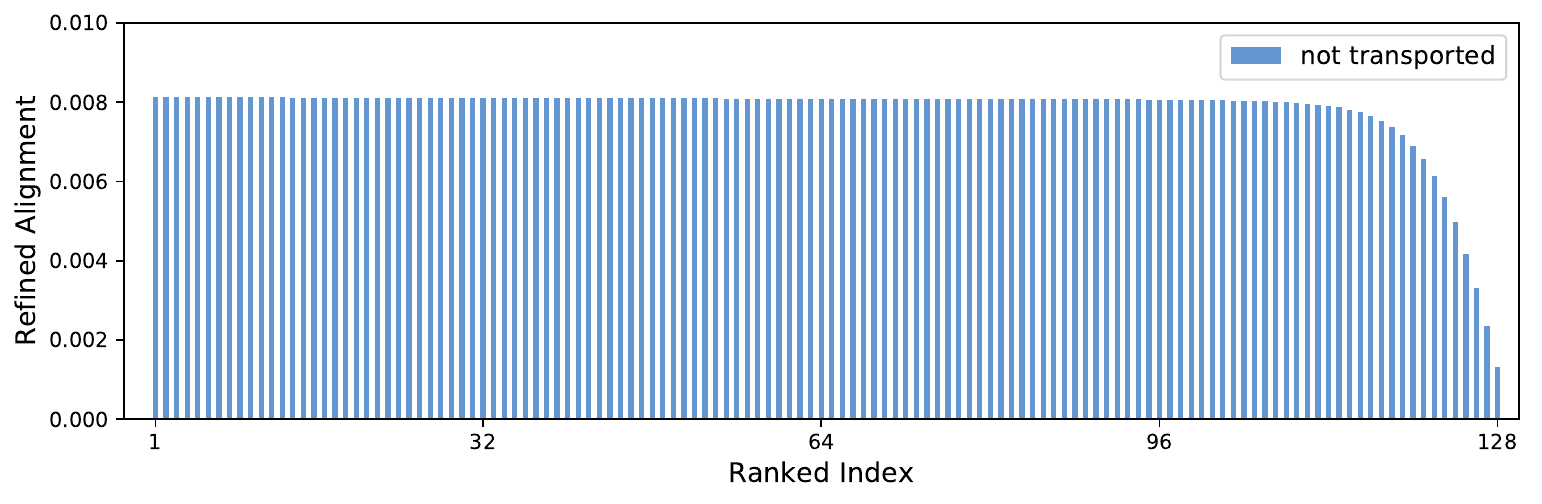}
	\caption{The averaged ranked distribution of normalized refined alignments (image to caption) about transported (upper subplot) and untransported (down subplot) data on MS-COCO under 0.4 PMPs. }
	\label{fig:t_and_ut}
\end{figure*}

\subsection{Analysis on Refined Alignment}
Our refined alignment is derived from a partial OT problem, which only allows $\rho$ unit mass to be transported. We further analyze how the transported and untransported data can benefit robust cross-modal retrieval. In \cref{fig:t_and_ut}, we plot the distribution of averaged refined alignments (image to caption) for both transported and untransported data drawn from each batch of the MS-COCO training set. The normalized distribution is ranked in descending order of probability. The upper subplot shows that the probability of transport data tends to concentrate on one dominant target. It is in line with our expectations that L2RM captures the semantic similarity among some unpaired samples. Interestingly, for those untransported data, the down subplot shows that the distribution of averaged refined alignments approximates a uniform distribution. Such refined alignments are formally equivalent to the label smoothing strategy, wherein the original one-hot targets are mixed with uniform target vectors, \ie{},
\begin{equation}\label{eq:label_smooth}
\bm{y}_i^{LS} = (1 - \gamma)\bm{y}_i + \frac{\gamma}{N_b - 1} (\mathbbm{1}_{N_b} - \bm{y}_i),
\end{equation}
where $\gamma$ is a smoothing parameter. As the original targets provide incorrect supervision for those mismatched pairs, increasing the value of $\gamma$ as much as possible can alleviate the impact of the wrong matching relation. Our refined alignments accord with this rule, which reveals that the untransported data can also improve the robustness against mismatched pairs.




\end{document}